\documentclass[10pt,twocolumn,letterpaper]{article}

\usepackage{cvpr}              

\usepackage{graphicx}
\usepackage{amsmath}
\usepackage{amssymb}
\usepackage{booktabs}

\usepackage[pagebackref,breaklinks,colorlinks]{hyperref}

\usepackage[capitalize]{cleveref}
\crefname{section}{Sec.}{Secs.}
\Crefname{section}{Section}{Sections}
\Crefname{table}{Table}{Tables}
\crefname{table}{Tab.}{Tabs.}

\def\cvprPaperID{6357} 
\def\confName{CVPR}
\def\confYear{2022}

\input{0.0-commands}

\begin{document}

\title{Towards Better Understanding Attribution Methods}
\author{Sukrut Rao, Moritz Böhle, Bernt Schiele\\
Max Planck Institute for Informatics, Saarland Informatics Campus, Saarbrücken, Germany\\
{\tt\small \{sukrut.rao,mboehle,schiele\}@mpi-inf.mpg.de}
}

\twocolumn[{%
\renewcommand\twocolumn[1][]{#1}%
\maketitle
\begin{center}
\begin{adjustbox}{minipage=\linewidth, scale=.93}
    \centering
    \vspace{-.8em}
    \captionsetup{type=figure}
    \includegraphics[width=\linewidth]{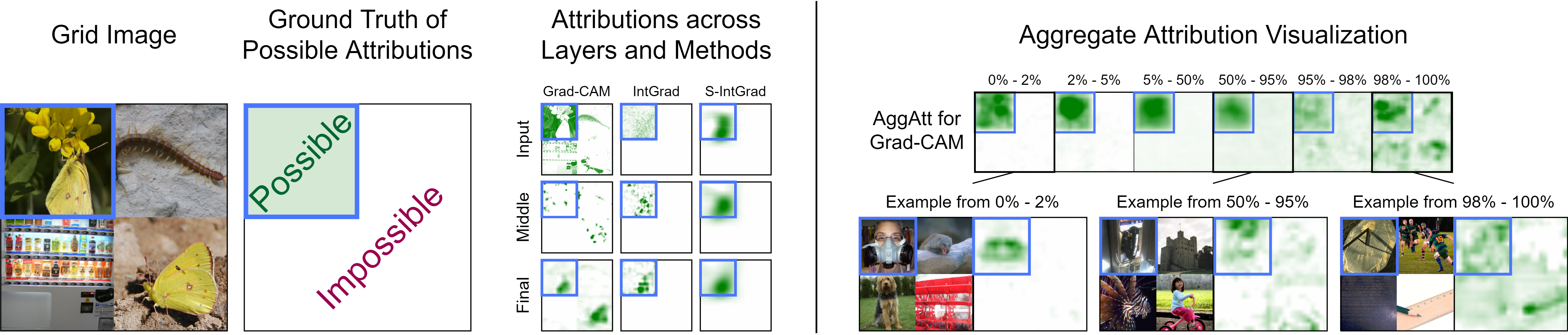}
\caption{
\textbf{Left:} Illustration of \emph{\disc} and \emph{ML-Att}. In \disc, we evaluate models on image grids (col.~1). Crucially, we employ separate \emph{classification heads} for each subimage that cannot possibly be influenced by other subimages; this yields `ground truths' for possible and impossible attributions (col.~2). For ML-Att, we evaluate methods at different network layers; here we show results for \gradcam and \intgrad. Further, we show results after smoothing \intgrad (\sintgrad), which we find to perform well (\cref{subsec:smoothing}).
\textbf{Right:} Visualisation of our \aggatt evaluation. By sorting attributions into percentile ranges w.r.t.~their performance and aggregating them over many samples, we obtain a holistic view of a methods' performance. AggAtt can thus reflect both best and worst case behaviour of an attribution method.
}
\label{fig:teaser}
    \vspace{-.25em}
\end{adjustbox}
\end{center}%
}]

\begin{abstract}
Deep neural networks are very successful on many vision tasks, but 
hard to interpret due to their black box nature. To overcome this, various
post-hoc attribution methods have been proposed to identify image regions most influential to the models' decisions. 
Evaluating such methods is challenging since no ground truth attributions exist. 
We thus propose three novel evaluation schemes
to more reliably measure the \emph{faithfulness} of those methods, to make comparisons between them more \emph{fair}, and to make \emph{visual inspection} more \emph{systematic}. 
To address faithfulness, we propose a novel evaluation setting (\difull) in which we carefully control which parts of the input can influence the output in order to distinguish possible from impossible attributions.
To address fairness, we note that different methods are applied at different layers, which skews any comparison,
and so evaluate all methods on the same layers (ML-Att) and discuss how this impacts their performance on quantitative metrics.
For more systematic visualizations, we propose a scheme (AggAtt) to qualitatively evaluate the methods on complete datasets.
We use these evaluation schemes to study strengths and shortcomings of some widely used attribution methods.
Finally, we propose a post-processing smoothing step that significantly improves the performance of some attribution methods,
and discuss its applicability.
\end{abstract}

\section{Introduction}
\label{sec:intro}

Deep neural networks (DNNs) are highly successful on many computer vision tasks.
However, their black box nature makes it hard to interpret and thus trust their decisions. 
To shed light on the models' decision-making process, several methods have been proposed that aim to attribute importance values to individual input features (see \cref{sec:related}).
However, given the lack of ground truth importance values, it has proven difficult to compare and evaluate these \emph{attribution methods} in a holistic and systematic manner.

In this work, we take a three-pronged approach towards addressing this issue. In particular, we focus on three important components for such evaluations: reliably measuring the methods' \emph{model-faithfulness}, ensuring a \emph{fair comparison} between methods, and providing a framework that allows for \emph{systematic} visual inspections of their attributions. 

First, we propose an evaluation scheme (\textbf{DiFull}), which allows distinguishing possible from impossible importance attributions. This effectively provides ground truth annotations for whether or not an input feature can possibly have influenced the model output. As such, it can highlight distinct failure modes of attribution methods (\cref{fig:teaser}, left). 

Second, a fair evaluation requires attribution methods to be compared on equal footing. However, we observe that different methods explain DNNs to different depths (e.g., full network or classification head only). 
Thus, some methods in fact solve a much easier problem (i.e., explain a much shallower network). To even the playing field, we propose a multi-layer evaluation scheme for attributions ({\bf ML-Att}) and provide a thorough evaluation of commonly used methods across multiple layers and models (\cref{fig:teaser}, left). When compared on the same level, we find that performance differences between some methods essentially vanish.

Third, relying on individual examples for a qualitative comparison is prone to skew the comparison and cannot fully represent the evaluated attribution methods. To overcome this, we propose a qualitative evaluation scheme for which we aggregate attribution maps ({\bf AggAtt}) across many input samples. This allows us to observe trends in the performance of attribution methods across complete datasets, in addition to looking at individual examples (\cref{fig:teaser}, right).

\myparagraph{Contributions.}
\textbf{(1)} We propose a novel evaluation setting, {\bf\disc}, 
in which we control which regions \emph{cannot possibly} influence a model's output, which allows us to highlight definite failure modes of attribution methods.
\textbf{(2)} 
We argue that methods can only be compared fairly when evaluated \emph{on the same layer}. To do this, we introduce {\bf ML-Att} and evaluate all attribution methods at multiple layers. 
We show that, when compared fairly, apparent performance differences between some methods effectively vanish. 
\textbf{(3)} We propose a novel aggregation method, {\bf\aggatt}, to qualitatively evaluate attribution methods across all images in a dataset. This allows to qualitatively assess a method's performance across many samples (\cref{fig:teaser}, right), which complements the evaluation on individual samples.
\textbf{(4)} We propose a post-processing smoothing step that significantly improves localization performance on some attribution methods. 
We observe significant differences when evaluating these smoothed attributions on different architectures, which highlights how architectural design choices can influence an attribution method's applicability. Our code is available at \href{https://github.com/sukrutrao/Attribution-Evaluation}{https://github.com/sukrutrao/Attribution-Evaluation}.

\section{Related Work}
\label{sec:related}

\myparagraph{Post-hoc attribution methods} 
broadly use one of three main mechanisms. 
\emph{Backpropagation-based} methods \cite{simonyan2013deep,springenberg2014striving,shrikumar2017learning,sundararajan2017axiomatic,srinivas2019full,zhang2018top} typically rely on the gradients with respect to the input~\cite{simonyan2013deep,shrikumar2017learning,sundararajan2017axiomatic,springenberg2014striving} or with respect to intermediate layers\cite{rebuffi2020there,zhang2018top}.
\emph{Activation-based} methods \cite{zhou2016learning,selvaraju2017grad,chattopadhyay2018grad,ramaswamy2020ablation,wang2020score,jiang2021layercam} weigh activation maps to assign importance, typically of the final convolutional layer.
The activations may be weighted by their gradients\cite{zhou2016learning,selvaraju2017grad,chattopadhyay2018grad,jiang2021layercam} or by estimating their importance to the classification score\cite{ramaswamy2020ablation,wang2020score}. 
\emph{Perturbation-based} methods \cite{ribeiro2016should,petsiuk2018rise,zeiler2013visualizing,fong2017interpretable,dabkowski2017real} treat the network as a black-box and assign importance by observing the change in output on perturbing the input. This is done by occluding parts of the image \cite{ribeiro2016should,petsiuk2018rise,zeiler2013visualizing} or 
optimizing for a mask that maximizes/minimizes class confidence\cite{fong2017interpretable,dabkowski2017real}.

In this work, we evaluate on a diverse set of attribution methods spanning all three categories.

\myparagraph{Evaluation Metrics:}
Several metrics have been proposed to evaluate attribution methods, and can broadly be categorised into \emph{Sanity checks}, 
\emph{localization-}, and \emph{perturbation-based metrics}. 
Sanity checks \cite{adebayo2018sanity,rebuffi2020there,ancona2017towards} test for basic properties an attribution method must satisfy (e.g., the explanation should depend on the model parameters). 
Localization-based metrics evaluate how well attributions localize class discriminative features of the input. 
Typically, this is done by measuring how well attributions coincide with object bounding boxes or image grid cells (see below)~\cite{zhang2018top,schulz2020restricting,cao2015look,fong2017interpretable,boehle2021convolutional}.
Perturbation-based metrics measure model behaviour under input perturbation to estimate feature importance.
Examples include removing the most\cite{samek2016evaluating} or least\cite{srinivas2019full} salient pixels, or using the attributions to scale input features and measuring changes in confidence\cite{chattopadhyay2018grad}.
Our work combines aspects from localization metrics and sanity checks to evaluate the model-faithfulness of an attribution method.

\myparagraph{Localization on Grids:}
Relying on object bounding boxes for localization assumes that the model only relies on information \emph{within} those bounding boxes.
However, neural networks are known to also rely on context information for their decisions, cf.~\cite{rakshith}.
Therefore, recent work \cite{boehle2021convolutional,shah2021input,arias2021explains} proposes creating a grid of inputs from distinct classes and measuring localization to the entire grid cell, which allows evaluation on datasets where bounding boxes are not available.
However, this does not \emph{guarantee} that the model only uses information from within the grid cell, and may fail for similar looking features (\cref{fig:vis:examplesmethods}, right). In our work, we propose a metric that controls the flow of information and guarantees that grid cells are classified independently.
\section{Evaluating Attribution Methods}
\label{sec:method}

\begin{figure*}[t]
\centering
\vspace{-.25em}
    \begin{subfigure}{0.24\linewidth}
        \centering
        \includegraphics[width=0.9\linewidth]{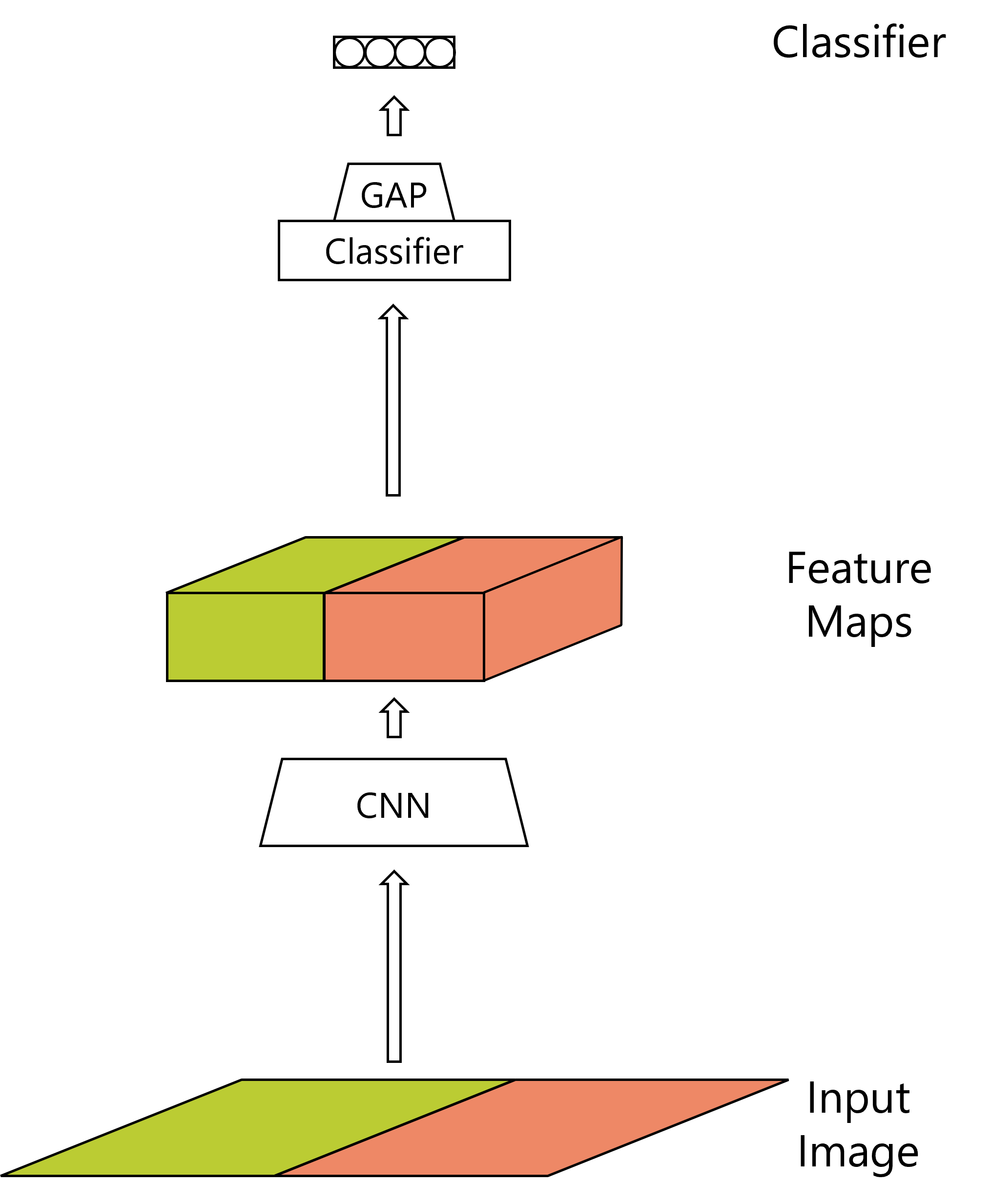}
        \caption{\pg}
        \label{fig:arch:pg}
    \end{subfigure}
    \rulesep
    \begin{subfigure}{0.31\linewidth}
        \centering
        \includegraphics[width=0.9\linewidth]{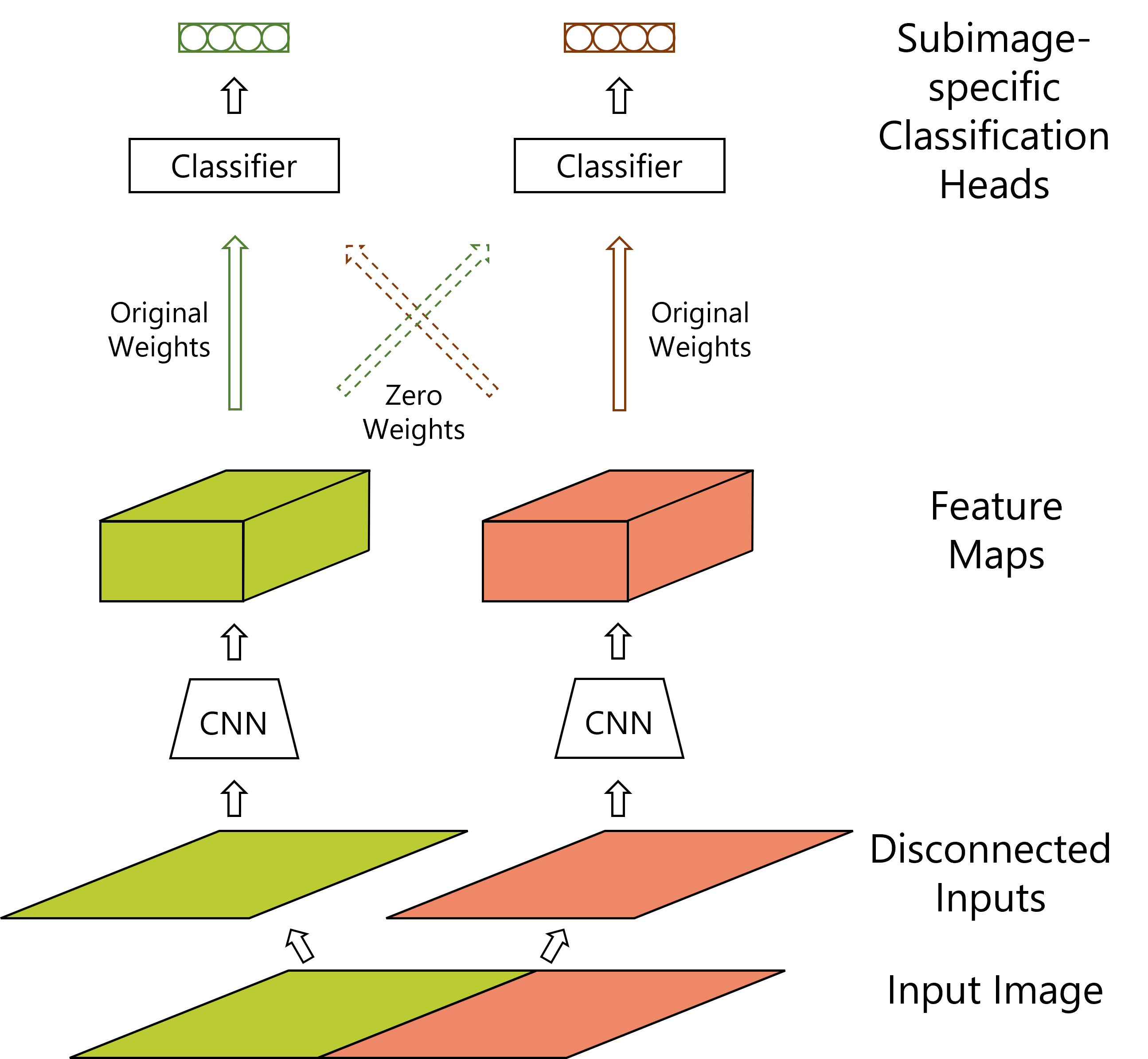}
        \caption{\disc}
        \label{fig:arch:disc}
    \end{subfigure}
    \rulesep
    \begin{subfigure}{0.27\linewidth}
        \centering
        \includegraphics[width=0.9\linewidth]{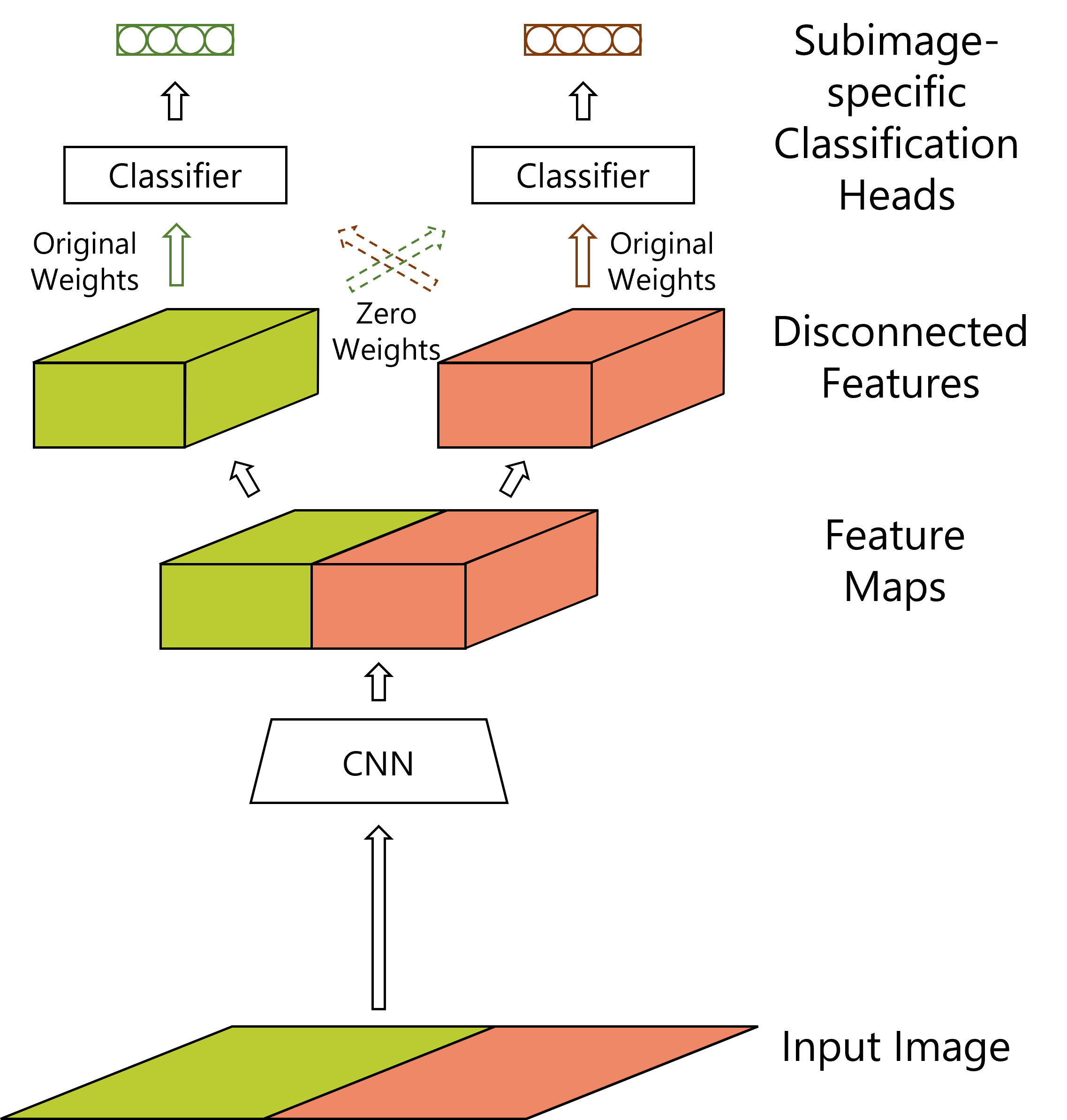}
        \caption{\cs}
        \label{fig:arch:cs}
    \end{subfigure}
    \caption{
    \textbf{Our three evaluation settings.} In \pg, the classification scores are influenced by the entire input. In \disc, on the other hand, we explicitly control which inputs can influence the classification score. For this, we pass each subimage separately through the spatial layers, and then construct individual classification heads for each of the subimages. \cs serves as a more natural setting to \disc, that still provides partial control over information. We show a $1\times 2$ grid for readability, but the experiments use $2\times 2$ grids.
    }
    \label{fig:arch}
    \vspace{-.85em}
\end{figure*}

We present our evaluation settings for better understanding the strengths and shortcomings of attribution methods. Similar to the Grid Pointing Game (\gridpg)\cite{boehle2021convolutional}, these metrics evaluate attribution methods on image grids with multiple classes. In particular, we propose a novel quantitative metric, \difull, and extension to it, \dipart (\ref{sec:method:quantitative}), as stricter tests of model faithfulness than \gridpg. Further, we present a qualitative metric, \aggatt (\ref{sec:method:aggatt}) and an evaluation setting that compares methods at identical layers, \mlatt (\ref{sec:method:intermediate}).

\subsection{Quantitative Evaluation: Disconnecting Inputs}\label{sec:method:quantitative}
In the following, we introduce the quantitative metrics that we use to compare attribution methods. For this, we first describe \pg and the grid dataset construction it uses\cite{boehle2021convolutional}. We then devise a novel setting, in which we carefully control which features can influence the model output. 
By construction, this provides ground truth annotations for image regions that can or cannot possibly have influenced the model output. While \pg evaluates how well the methods localize class discriminative features, our metrics complement it by evaluating their \emph{model-faithfulness}.

\mysubsub{Grid Data and \pg}\label{sec:method:gridpg}
For \pg~\cite{boehle2021convolutional}, the attribution methods are evaluated on a synthetic grid of $n\times n$ images in which each class may occur at most once. In particular, for each of the occurring classes, \pg measures the fraction of positive attribution assigned to the respective grid cell versus the overall amount of positive attribution. Specifically, let $A^+(p)$ refer to the positive attribution given to the $p^{th}$ pixel. The localization score for the subimage $x_i$ is given by:
\begin{equation}
    \label{eq:localisation}
    L_i = \frac{\sum_{p\in x_i} A^+(p)}{\sum_{j=1}^{n^2}\sum_{p\in x_j} A^+(p)} 
\end{equation}
An `optimal' attribution map would thus yield $L_i\myeq1$, while uniformly distributing attributions would yield $L_i\myeq\frac{1}{n^2}$.
 
 By only using confidently classified images from distinct classes, \pg aims to ensure that the model does not find `positive evidence' for any of the occurring classes in the grid cells of other classes. 
 However, specifically for class-combinations that share low-level features, this assumption might not hold, see \cref{fig:vis:examplesmethods} (right): despite the two dogs (upper left and lower right) being classified correctly as single images, the output for the logit of the dog in the upper left is influenced by the features of the dog in the lower right in the grid image.
 Since all images in the grid can indeed influence the model output in \pg\footnote{As shown in \cref{fig:arch:pg}, the convolutional layers of the model under consideration process the entire grid to obtain feature maps, which are then classified point-wise. Finally, a \emph{single output per class} is obtained by globally pooling all point-wise classification scores. As such, the class logits can, of course, be influenced by all images in the grid.}, it is unclear whether such an attribution is in fact not \emph{model-faithful}.
 
\mysubsub{Proposed Metric: \disc}\label{sec:method:difull}
As discussed, the assumption in \pg that no feature outside the subimage of a given class should positively influence the respective class logit might not hold. Hence, we propose to \underline{full}y \underline{di}sconnect (\disc) the individual subimages from the model outputs for other classes. 
For this, we introduce two modifications. First, after removing the GAP operation, we use $n \times n$ classification heads, one for each subimage, only \emph{locally} pooling those outputs that have their receptive field center \emph{above the same subimage}. Second, we ensure that their receptive field {does not overlap} with other subimages by zeroing out the respective connections. 

In particular, we implement \disc by passing the subimages separately through the CNN backbone of the model under consideration\footnote{Note that this is equivalent to setting the respective weights of a convolutional kernel to zero every time it overlaps with another subimage.}, see \cref{fig:arch:disc}. Then, we apply the classification head separately to the feature maps of each subimage. As we discuss in the supplement, \difull has similar computational requirements as \gridpg.

As a result, we can \emph{guarantee} that no feature outside the subimage of a given class can possibly have influenced the respective class logit---they are indeed \emph{fully disconnected}.

\mysubsub{Natural Extension: \cs}\label{sec:method:dipart}
At one end, \pg allows any subimage to influence the output for any other class, while at the other, \disc completely disconnects the subimages. In contrast to \pg, \disc might be seen as a constructed setting not seen in typical networks. 
As a more natural setting, we therefore propose \cs, for which we only \underline{part}ially \underline{di}sconnect the subimages from the outputs for other classes, see \cref{fig:arch:cs}.  Specifically, we do not zero out all connections (\cref{sec:method:difull}), but instead only apply the local pooling operation from \disc and thus obtain local classification heads for each subimage (as in \disc). 
However, in this setting, the classification head for a specific subimage can be influenced by features in other subimages that lie within the head's receptive field. For models with a small receptive field, this yields very similar results as \disc (\cref{sec:results} and Supplement).

\subsection{Qualitative Evaluation: \aggatt}\label{sec:method:aggatt}
In addition to quantitative metrics, attribution methods are often compared qualitatively on individual examples for a visual assessment. However, this is sensitive to the choice of examples and does not provide a holistic view of the method's performance.
By constructing standardised grids, in which `good' and `bad' (\pg) or \emph{possible} and \emph{impossible} (\disc) attributions are always located in the same regions, we can instead construct \emph{aggregate attribution maps}. 

Thus, we propose a new qualitative evaluation scheme, \mbox{\bf\aggatt}, for which we generate a set of aggregate maps for each method that progressively show the performance of the methods from the best to the worst localized attributions. 

For this, we first select a grid location and then sort all corresponding attribution maps in descending order of the localization score, see \cref{eq:localisation}. Then, we bin the maps into percentile ranges and, finally, obtain an aggregate map per bin by averaging all maps within a single bin. In our experiments, we observed that attribution methods typically performed consistently over a wide range of inputs, but showed significant deviations in the tails of the distributions (best and worst case examples). Therefore, to obtain a succinct visualization that highlights both distinct failure cases as well as the best possible results, we use bins of unequal sizes. Specifically, we use smaller bins for the top and bottom percentiles. For an example of \aggatt, see \cref{fig:teaser}.

As a result, \aggatt allows for a \emph{systematic} qualitative evaluation and provides a holistic view of the performance of attribution methods across many samples.

\subsection{Attributions Across Network Layers: ML-Att}\label{sec:method:intermediate}
Attribution methods often vary significantly in the degree to which they explain a model. Activation-based attribution methods like \gradcam\cite{selvaraju2017grad}, e.g., are typically applied on the last spatial layer, and thus only explain a fraction of the full network.
This is a significantly easier task as compared to explaining the entire network, as is done by typical backpropagation-based methods. Activations from deeper layers of the network would also be expected to localize better, since they would represent the detection of higher level features by the network (\cref{fig:teaser}, left).

For a \emph{fair comparison} between methods, we thus propose a multi-layer evaluation scheme for attributions ({\bf ML-Att}). Specifically, we evaluate methods at various network layers and compare their performance \emph{on the same layers}. For this, we evaluate all methods at the input, an intermediate, and the final spatial layer of multiple network architectures, see \cref{sec:experiments} for details.
Importantly, we find that apparent differences found between some attribution methods vanish when compared \emph{fairly}, i.e., on the same layer (\cref{subsec:layer_results}).

Lastly, we note that most attribution methods have been designed to assign importance values to \emph{input features} of the model, not \emph{intermediate} network activations. The generalisation to intermediate layers, however, is straightforward. For this, we simply divide the full model $\mathbf f_\text{full}$ into two virtual parts: $\mathbf f_\text{full}\myeq\mathbf f_\text{explain}\circ \mathbf f_\text{pre}$. Specifically, we treat $\mathbf f_\text{pre}$ as a pre-processing step and use the attribution methods to explain the outputs of $\mathbf f_\text{explain}$ with respect to the inputs $\mathbf f_\text{pre}(\mathbf x)$. Note that in its standard use case, in \gradcam $\mathbf f_\text{pre}(\mathbf x)$ is given by all convolutional layers of the model, whereas for most gradient-based methods  $\mathbf f_\text{pre}(\mathbf x)$ is the identity.
\section{Experimental Setup}
\label{sec:experiments}

\begin{figure*}[!t]
\centering
\includegraphics[width=\linewidth]{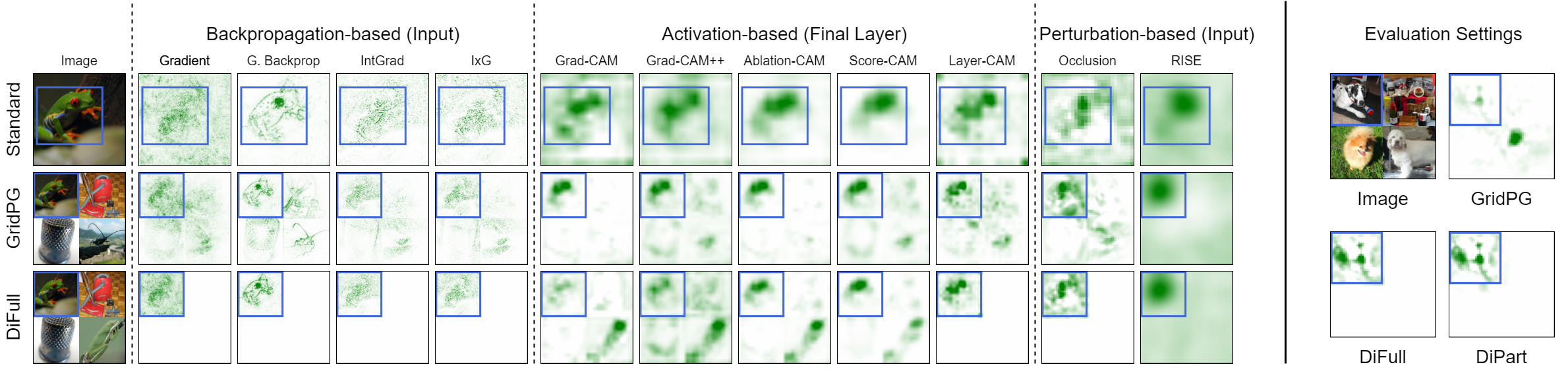}
\caption{\textbf{Left:} Example Attributions on the Standard, \pg, and \disc Settings. We show attributions for all methods on their typically evaluated layers, \ie input for backpropagation-based and perturbation-based, and final layer for activation-based methods. Blue boxes denote the object bounding box (Standard) or the grid cell (\pg, \disc) respectively. For \disc, we use images of the same class at the top-left and bottom-right corners as in our experiments. \textbf{Right:} \occ attributions for an example evaluated on \pg, \disc, and \cs. The top-left and bottom-right corners contain two different species of dogs, which share similar low-level features, causing both to be attributed in \pg. In contrast, our disconnected construction in \disc and \cs ensures that the bottom-right subimage does not influence the classification of the top-left, and thus should not be attributed by any attribution methods, even though some do erroneously.}
\label{fig:vis:examplesmethods}
\end{figure*}

\myparagraph{Dataset and Architectures:}
We run our experiments on VGG11\cite{simonyan2014very} and Resnet18\cite{he2016deep} trained on Imagenet\cite{imagenet}; similar results were observed on CIFAR10\cite{krizhevsky2009learning} (in supplement).  For each model, we separately select images from the validation set that were classified with a confidence score of at least 0.99. By only using highly confidently classified images\cite{boehle2021convolutional,arias2021explains}, we ensure that the features within each grid cell constitute positive evidence of its class for the model, and features outside it contain low positive evidence since they get confidently classified to a different class.

\myparagraph{Evaluation on \pg, \disc, and \cs:}
We evaluate on $2\mytimes2$ grids constructed by randomly sampling images from the set of confidently classified images (see above). Specifically, we generate 2000 attributions per method for each of \pg, \disc, and \cs. For \pg, we use images from distinct classes, while for \disc and \cs we use distinct classes except in the bottom right corner, where we use the same class as the top left. By repeating the same class twice, we can test whether an attribution method simply highlights class-related features, irrespective of them being used by the model.
Since subimages are disconnected from the classification heads of other locations in \disc and \cs, the use of repeating classes does not change which regions should be attributed (\cref{sec:method:difull}).

\myparagraph{Evaluation at Intermediate Layers:}
We evaluate each method at the input (image), middle\footnote{We show a single intermediate layer to visualize trends from the input to the final layer. Results on all layers can be found in the supplement.} (Conv5 for VGG11, Conv3\_x for Resnet18), and final spatial layer (Conv8 for VGG11, Conv5\_x for Resnet18) of each network, see \cref{sec:method:intermediate}. Evaluating beyond the input layer leads to lower dimensional attribution maps, given by the dimensions of the activation maps at those layers. 
Thus, as is common practice~\cite{selvaraju2017grad}, we upsample those maps to the dimensions of the image ($448\times 448$) using bilinear interpolation.

\myparagraph{Qualitative Evaluation on \aggatt:}
As discussed, for \aggatt we use bins of unequal sizes (\cref{sec:method:aggatt}). In particular, we bin the attribution maps into the following percentile ranges: 0--2\%, 2--5\%, 5--50\%, 50--95\%, 95--98\%, and 98--100\%; cf.~\cref{fig:teaser}. Further, in our experiments we evaluate the attributions for classes at the top-left grid location.

\myparagraph{Attribution Methods:}
We evaluate a diverse set of attribution methods, for an overview see \cref{sec:related}. As discussed in \cref{sec:method:intermediate}, to apply those methods to intermediate network layers, we divide the full model into two virtual parts $\mathbf f_\text{pre}$ and $\mathbf f_\text{explain}$ and treat the output of  $\mathbf f_\text{pre}$ as the input to $\mathbf f_\text{explain}$ to obtain importance attributions for those `pre-processed' inputs. In particular, we evaluate the following methods. 
From the set of \textbf{backpropagation-based methods}, we evaluate on \gb\cite{springenberg2014striving}, \grad\cite{simonyan2013deep}, \intgrad\cite{sundararajan2017axiomatic}, and \ixg\cite{shrikumar2017learning}. 
From the set of \textbf{activation-based methods}, we evaluate on \gradcam\cite{selvaraju2017grad}, \gradcampp\cite{chattopadhyay2018grad}, \ablationcam\cite{ramaswamy2020ablation}, \scorecam\cite{wang2020score}, and \layercam\cite{jiang2021layercam}. 
Note that in our framework, these methods can be regarded as using the classification head only (except \cite{jiang2021layercam}) for $\mathbf f_\text{explain}$, see \cref{sec:method:intermediate}. In order to evaluate them at earlier layers, we simply expand $\mathbf f_\text{explain}$ accordingly to include more network layers.
From the set of \textbf{perturbation-based methods}, we evaluate \occ\cite{zeiler2013visualizing} and \rise\cite{petsiuk2018rise}. These are typically evaluated on the input layer, and measure output changes when perturbing (occluding) the input (\cref{fig:vis:examplesmethods}, left). 
Note that \occ involves sliding an occlusion kernel of size $K$ with stride $s$ over the input.
We use $K\myeq16,s\myeq8$ for the input, and $K\myeq5,s\myeq2$ at the middle and final layers to account for the lower dimensionality of the feature maps.
For \rise, we use $M\myeq1000$ random masks, generated separately for evaluations at different network layers.

\section{Experimental Results and Discussion}
\label{sec:results}

\begin{figure*}[t]
\includegraphics[width=\linewidth]{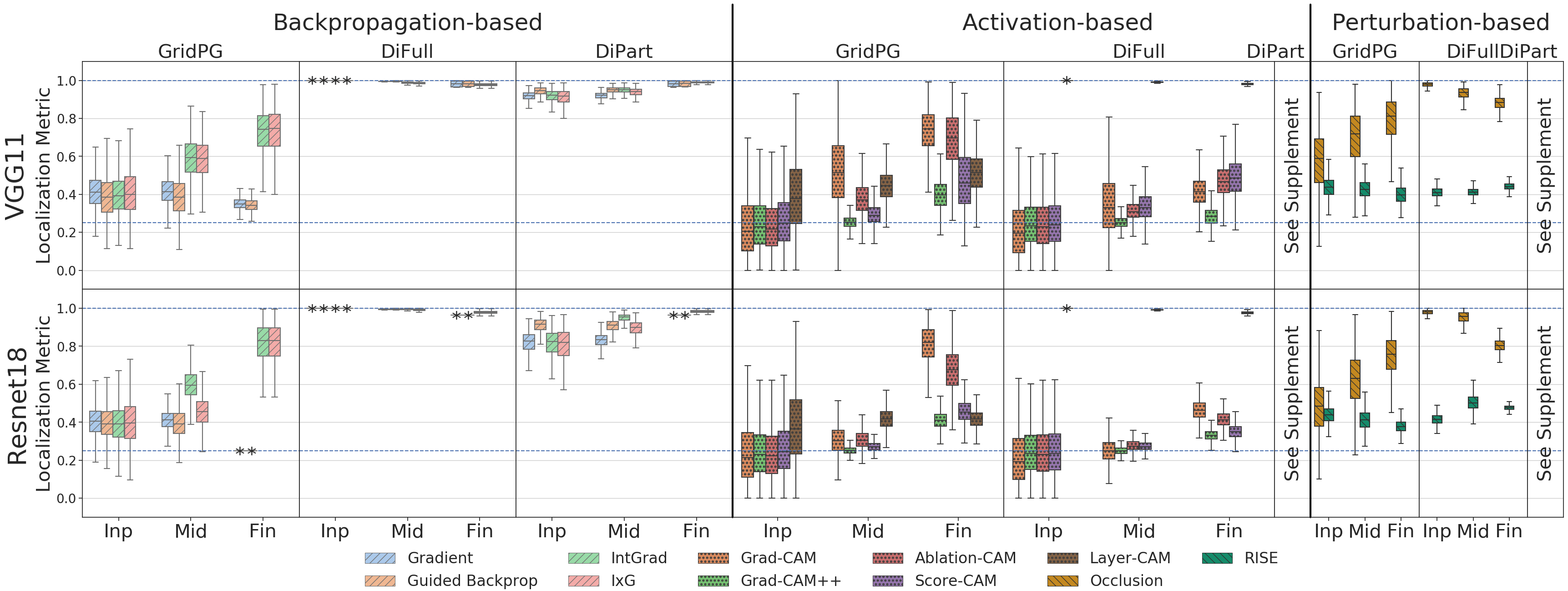}
\caption{\textbf{Quantitative Results on VGG11 and Resnet18.} For each metric, we evaluate all attribution methods with respect to the input image (Inp), a middle (Mid), and the final (Fin) spatial layer. We observe the performance to improve from \emph{Inp} to \emph{Fin} on most settings. Similar to the backpropagation-based methods, the results for methods on \cs are very similar to those in \disc for activation and perturbation-based; for details, see supplement.
The symbol * denotes boxes that collapse to a single value, for better readability.
}
\label{fig:main}
\end{figure*}
\begin{figure}[!t]
\includegraphics[width=\linewidth]{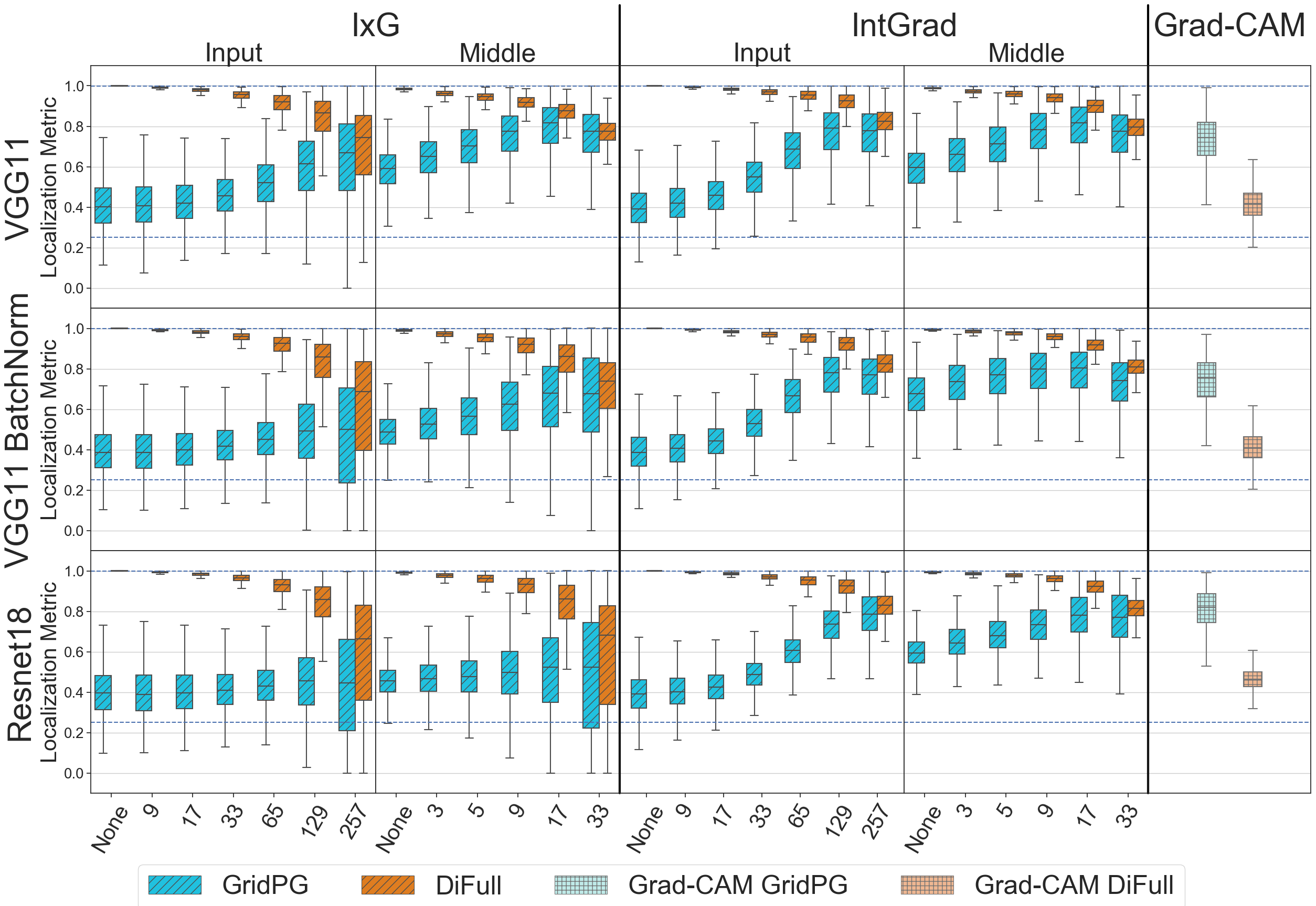}
\caption{\textbf{Smoothing the attributions} for \intgrad and \ixg significantly improves their performance at the input image and middle layer. For reference, we show \gradcam on the \emph{final} spatial layer.
}
\label{fig:smooth}
\end{figure}
\begin{figure*}[t]
\vspace{-.5em}
\includegraphics[width=\linewidth]{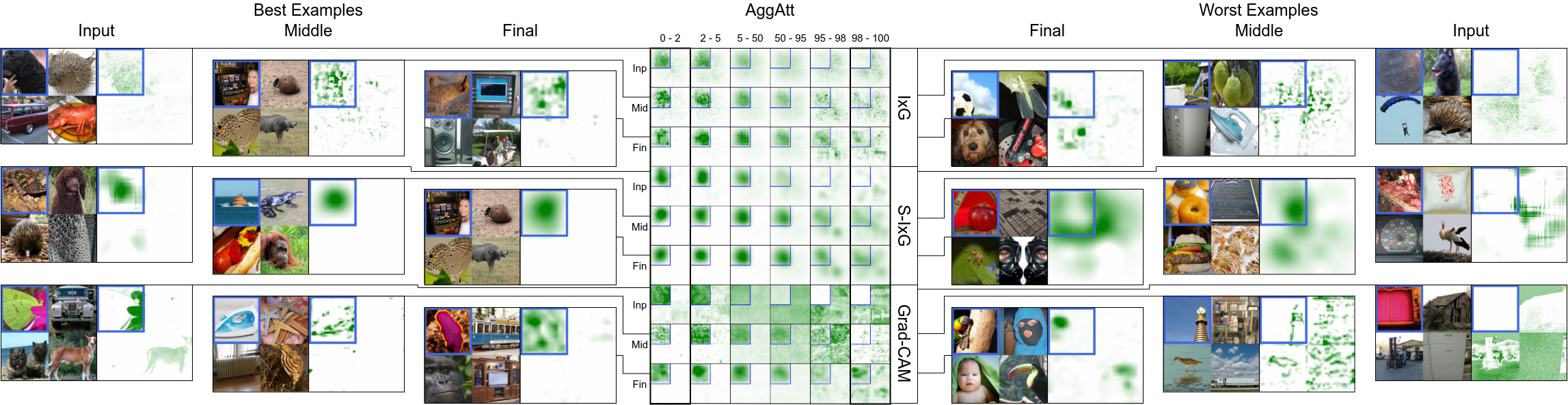}
\caption{\textbf{Qualitative Results for VGG11 on \pg evaluated at the top-left corner}. \textit{Centre:} Aggregate attributions sorted and binned in descending order of localization. Each column corresponds to a bin, and set of three rows corresponds to a method. For each method, the three rows from top to bottom show the aggregate attributions at the input, middle, and final spatial layers. \textit{Left:} Examples from the first bin, which corresponds to the best set of attributions. \textit{Right:} Similarly, we show examples from the last bin, which corresponds to the worst set of attributions. For smooth \ixg, we use $K=129$ for the input layer, $K=17$ at the middle layer, and $K=9$ at the final layer. All examples shown correspond to images whose attributions lie at the median position in their respective bins.
}
\vspace{-.5em}
\label{fig:vis:pg}
\end{figure*}
\begin{figure}[t]
\includegraphics[width=\linewidth]{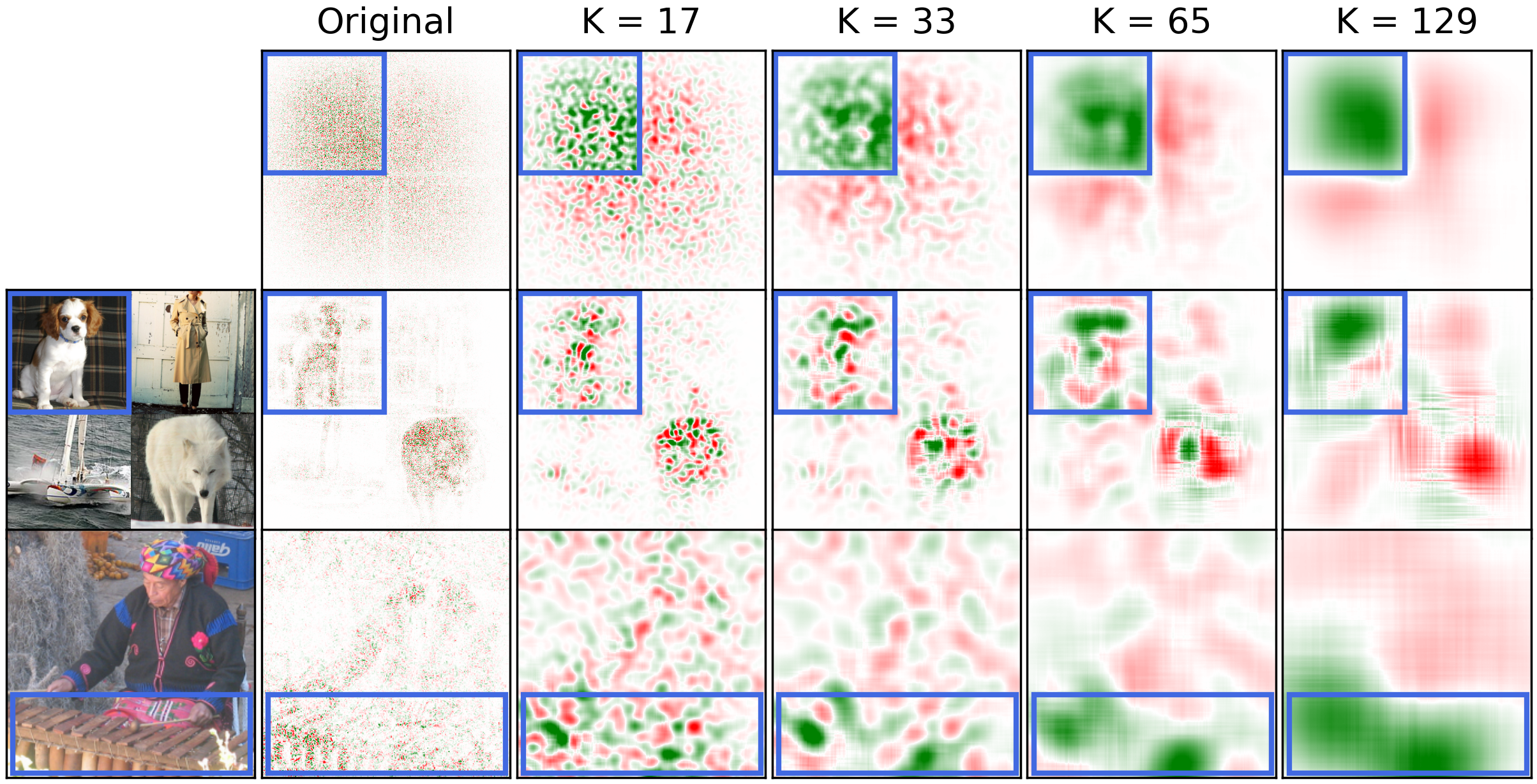}
\caption{\textbf{Qualitative Visualization of smoothing \ixg attribution maps for various kernel sizes, including both positive and negative attributions.} \emph{Top}: Aggregate attribution maps for VGG11 on \pg at the top-left corner across the dataset. We see that positive attributions (green) aggregate to the top-left grid cell and negative attributions (red) aggregate outside when smoothing with large kernel sizes. \emph{Middle and Bottom}: Examples of smoothing on a single grid and non-grid image. 
Positive attributions concentrate inside the bounding box when smoothed with large kernels.
}
\label{fig:vis:ixgsmoothnegative}
\end{figure}
\begin{figure*}[t]
\includegraphics[width=\linewidth]{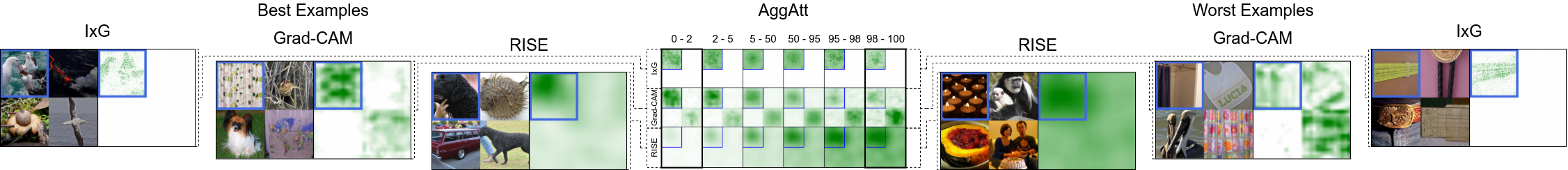}
\caption{\textbf{Qualitative Results for VGG11 on \disc evaluated at the top-left corner}. \textit{Centre:} Aggregate attributions sorted and binned in descending order of localization. Each column corresponds to a bin and each row corresponds to a method applied at its standard layer. \textit{Left:} Examples from the first bin, which corresponds to the best set of attributions. \textit{Right:} Examples from the last bin, which corresponds to the worst set of attributions. All examples shown correspond to images whose attributions lie at the median position in their bins.}
\label{fig:vis:disc}
\end{figure*}

In this section, we first present the quantitative results for all attribution methods on \pg, \cs, and \disc and compare their performance at multiple layers (\ref{subsec:layer_results}).
Further, we present a simple smoothing mechanism that provides highly performant attributions on all three settings, and discuss architectural considerations that impact its effectiveness (\ref{subsec:smoothing}). Finally, we present qualitative results using \aggatt, and show its use in highlighting strengths and deficiencies of attribution methods (\ref{subsec:evalqualitative}).

\subsection{Evaluation on \pg, \disc, and \cs}
\label{subsec:layer_results}
We perform \mlatt evaluation using the input (Inp), a middle layer (Mid), and before the classification head (Fin) (x-ticks in \cref{fig:main}) for all three quantitative evaluation settings (\pg, \disc, \cs, minor columns in \cref{fig:main}) discussed in \cref{sec:method}. In the following, we discuss the methods' results, grouped by their `method family': backpropagation-based, activation-based, and perturbation-based methods (major columns in \cref{fig:main}).

\myparagraph{Backpropagation-based methods:}
We observe that all methods perform poorly at the initial layer on \pg (\cref{fig:main}, left). 
Specifically, we observe gradient-based methods to yield noisy attributions that do not seem to reflect the grid structure of the images; i.e., positive attributions are nearly as likely to be found outside of a subimage for a specific class as they are to be found inside.

However, all methods improve on later layers. At the final layer, \intgrad and \ixg show very good localization (comparable to \gradcam), which suggests that the methods may have similar explanatory power when compared on an equal footing. We note that \ixg at the final layer has been previously proposed under the name \detgradcam\cite{saha2020role}.

On \disc, all methods show near-perfect localization across layers (\cref{fig:vis:disc}). No attribution is given to disconnected subimages since the gradients with respect to them are zero (after all, they are \emph{fully disconnected});
degradations for other layers can be attributed to the applied upsampling.

Similar results are seen in \cs, but with decreasing localization when moving backwards from the classifier, which can be attributed to the fact that the receptive field can overlap with other subimages in this setting.
Given the similarity between the results for \disc and \cs, we restrict our discussion to \disc; for \cs, see supplement.

\myparagraph{Activation-based methods:}
We see that all methods with the exception of \layercam improve in localization performance from input to final layer on all three settings. Since attributions are computed using a scalar weighted sum of attribution maps, this improvement could be explained by improved localization of activations from later layers. In particular, localization is very poor at early layers, which is a well-known limitation (cf. \cite{jiang2021layercam}) of \gradcam. The weighting scheme also causes final layer attributions for all methods except \layercam to perform worse on \disc than on \pg, since these methods attribute importance to both instances of the repeated class (\cref{fig:vis:disc}). This issue is absent in \layercam since it does not apply a pooling operation.

\myparagraph{Perturbation-based methods:}
We observe (\cref{fig:main}, right) \occ to perform well across layers on \disc, since occluding disconnected subimages cannot affect the model outputs and are thus not attributed importance. 
However, the localization drops slightly for later layers. This is due to the fact that the relative size (w.r.t.~activation map) of the overlap regions between occlusion kernels and adjacent subimages increases.
This highlights the sensitivity of performance to the choice of hyperparameters, and the tradeoff between computational cost and performance. 

On \pg, \occ performance improves with layers.
On the other hand, \rise performs poorly across all settings and layers. Since it uses random masks, pixels outside a target grid cell that share a mask with pixels within get attributed equally. So while attributions tend to concentrate more in the target grid cell, the performance can be inconsistent (\cref{fig:vis:disc}).

\subsection{Smoothing Attributions}\label{subsec:smoothing}
From \cref{subsec:layer_results}, we see that \gradcam localizes well at the final layer in \pg, but performs poorly on all the other settings as a consequence of global pooling of gradients (for \disc) and poor localization of early layer features (for \pg early layers). 
Since \ixg, in contrast, does not use a pooling operation, it performs well on \disc at all layers and on \pg at the final layer.
However, it performs poorly at the input and middle layers on \pg due to the noisiness of gradients; \intgrad shows similar results. 

Devising an approach to eliminate this noise would provide an attribution method that performs well across settings and layers. 
Previous approaches to reduce noise include averaging attribution maps over many perturbed samples (SmoothGrad\cite{smilkov2017smoothgrad}, see supplement for a comparison) or adding a gradient penalty during training\cite{kiritoshi2019l1}. However, \smoothgrad is computationally expensive as it requires several passes on the network to obtain attributions, and is sensitive to the chosen perturbations. 
Similarly, adding a penalty term during training requires retraining the network. 

Here, we propose to simply apply a Gaussian smoothing kernel on existing \intgrad and \ixg attributions. We evaluate on \disc and \pg using several kernel sizes, using standard deviation $K/4$ for kernels of size $K$. We refer to the smooth versions as \sintgrad and \sixg respectively.

On VGG11 (\cref{fig:smooth}, top), we find that \sintgrad and \sixg localize significantly better than \intgrad and \ixg, and the performance improves with increasing kernel size. In detail, \sintgrad \emph{on the input layer} with $K\myeq257$ outperforms \gradcam \emph{on the final layer}, despite explaining the full network. While performance on \disc drops slightly as smoothing leaks attributions across grid boundaries, both \sintgrad and \sixg localize well across settings and layers. However, on Resnet18 (\cref{fig:smooth}, bottom), while \sintgrad improves similarly, \sixg does not, which we discuss next.

\myparagraph{Impact of Network Architecture:}\label{subsec:batchnorm}
A key difference between the VGG11 and Resnet18 architectures used in our experiments is that VGG11 does not have batch normalization (BatchNorm) layers.
We note that batch norm effectively randomizes the sign of the input vectors to the subsequent layer, by centering those inputs around the origin (cf.\cite{ioffe2015batch,kiritoshi2019l1}). Since the sign of the input determines whether a contribution (weighted input) is \emph{positive} or \emph{negative}, a BatchNorm layer will randomize the sign of the contribution and the `valence' of the contributions will be encoded in the BatchNorm biases.
To test our hypothesis, we evaluate \sixg on a VGG11 with BatchNorm layers (\cref{fig:smooth}, middle), and observe results similar to Resnet18: i.e., we observe no systematic improvement by increasing the kernel size of the Gaussian smoothing operation. This shows that the architectural choices of a model can have a significant impact on the performance of attribution methods.

\subsection{Qualitative Evaluation using \aggatt}\label{subsec:evalqualitative}
In this section, we present qualitative results using \aggatt for select attributions evaluated on \pg and \disc and multiple layers. 
First, to investigate the qualitative impact of smoothing, we use \aggatt to compare \ixg, \sixg, and \gradcam attributions on \pg on multiple layers.
We employ \aggatt on \disc to highlight specific characteristics and failure cases of some attribution methods.

\myparagraph{\aggatt on \pg:}
We show \aggatt results for \ixg, \sixg, and \gradcam at three layers on \pg using VGG11 on the images at the top-left corner (\cref{fig:vis:pg}).
For each method, a set of three rows corresponds to the attributions at input, middle, and final layers. For \sixg, we set $K$ to $129$, $17$, and $9$ respectively.
We further show individual samples (median bin) of the first and last bins per method. 

We observe that the aggregate visualizations are consistent with the quantitative results (\cref{fig:main,,fig:smooth}) and the individual examples shown for each bin. 
The performance improves for \ixg and \gradcam from input to final layer, while \sixg localizes well across three layers. Finally, the last two columns show that all the attribution methods perform `poorly' for some inputs; e.g., we find that \ixg and \gradcam on the final layer attribute importance to other subimages if they exhibit features that are consistent with the class in the top-left subimage.
While the attributions might be conceived as incorrect, we find that many `failure cases' on \pg highlight features that the underlying model might in fact use, even if they are in another subimage. 
Given the lack of ground truth, it is difficult to assess whether these attributions faithfully reflect model behaviour or deficiencies of the attribution methods.
 
Despite explaining significantly more layers, \sintgrad and \sixg at the input layer not only match \gradcam at the final layer quantitatively (\cref{fig:smooth}) and qualitatively (\cref{fig:vis:pg}), but are also highly consistent with it for individual explanations. Specifically, the Spearman rank correlation between the localization scores of \gradcam (final layer) and \sintgrad (input layer) increases significantly as compared to \intgrad (input layer)  (e.g., $0.34$$\rightarrow$$0.80$ on VGG11), implying that their attributions for any input tend to lie in the same \aggatt bins (see supplement).

To further understand the effect of smoothing, we visualize \sixg with varying kernel sizes while including negative attributions (\cref{fig:vis:ixgsmoothnegative}). The top row shows aggregate attributions across the dataset, while the middle and bottom rows show an example under the \pg and standard localization settings respectively. We observe that while \ixg attributions appear noisy (column 2), smoothing causes positive and negative attributions to cleanly separate out, with the positive attributions concentrating around the object. For instance, in the second row, \ixg attributions concentrate around both the dog and the wolf, but \sixg with $K\myeq129$ correctly attributes only the dog positively. This could indicate a limited effective receptive field (RF) \cite{luo2016understanding} of the models. Specifically, note that for piece-wise linear models, summing the contributions (given by \ixg) over all input dimensions within the RF exactly yields the output logit (disregarding biases). Models with a small RF would thus be well summarised by \sixg for an adequately sized kernel; we elaborate on this in the supplement.

\myparagraph{\aggatt on \disc:}
We visually evaluate attributions on \disc for one method per method family, i.e., from backpropagation-based (\ixg, input layer), activation-based (\gradcam, final layer), and perturbation-based (\rise, input layer) methods at their standard layers (\cref{fig:vis:disc}). The top row corroborates the near-perfect localization shown by the backpropagation-based methods on \disc. The middle row shows that \gradcam attributions concentrate at the top-left and bottom-right corners, which contain images of the same class, since global pooling of gradients makes it unable to distinguish between the two even though only the top-left instance (here) influences classification. Finally, for \rise, we observe that while attributions localize well for around half the images, the use of random masks results in noisy attributions for the bottom half.

\section{Conclusion}
\label{sec:conclusion}

In this work, we proposed schemes to evaluate model-faithfulness of attribution methods in a fair and systematic manner. We first proposed a quantitative metric, \disc, that constrains input regions that can affect classification. This yields regions in which any attribution is necessarily unfaithful and allows to highlight distinct failure cases of some attribution methods.
Then, we proposed a multi-layer evaluation scheme, \mlatt, that compares methods fairly, and found that the performance gap between methods narrows considerably when done so. Finally, we proposed \aggatt, a novel qualitative evaluation scheme that allows for succinctly visualizing the variation in performance of attribution methods in a systematic and holistic manner. 
Overall, we find that fair comparisons, holistic evaluations (\difull, \gridpg, \aggatt, \mlatt), and careful disentanglement of model behaviour from the explanations provide better insights in the performance of attribution methods.

\myparagraph{Limitations.} We note that while our method can distinguish between possibly correct and impossibly correct attributions, our method cannot evaluate the correctness of specific attributions within the target grid cells, since ground truths for these are unknown.

{\small
\bibliographystyle{ieee_fullname}
\bibliography{references}
}

\end{document}


\title{Towards Better Understanding Attribution Methods\\Supplementary Material}

\author{Sukrut Rao, Moritz Böhle, Bernt Schiele\\
Max Planck Institute for Informatics, Saarland Informatics Campus, Saarbrücken, Germany\\
{\tt\small \{sukrut.rao,mboehle,schiele\}@mpi-inf.mpg.de}
}
\maketitle

\tableofcontents
~\\
In this supplement, we provide additional details and results of our evaluation on our proposed settings. First, we provide additional quantitative results on ImageNet \cite{imagenet} (\cref{sec:additional:dipart}), including complete results on \dipart, followed by complete qualitative results using \aggatt (\cref{sec:additional:aggatt}) with examples. Then, we investigate the correlation between the localization scores of attribution methods (\cref{sec:additional:correlation}) and compare it with the trends seen from the qualitative and quantitative results. This is followed by a discussion of the impact of smoothing attributions (\cref{sec:additional:smoothing}), including a comparison of the similarity of \gradcam\cite{selvaraju2017grad} with \sintgrad and \sixg on the same set of examples across \aggatt bins. In \cref{sec:alllayers}, we provide quantitative results at all spatial layers for both models, and show that it reflects the trends observed from the three chosen layers in our experiments. This is followed by an analysis of the computational costs of each evaluation setting (\cref{sec:cost}), and a comparison of our proposed smoothing technique with SmoothGrad \cite{smilkov2017smoothgrad} (\cref{sec:smoothgrad}). Then, we describe further implementation details (\cref{sec:impldetails}) of our experiments. Finally, we provide results of our approach on CIFAR10 \cite{krizhevsky2009learning} (\cref{sec:cifar}), and find similar trends as on ImageNet across our settings.

\newpage


\section{Quantitative Results Including \dipart}\label{sec:additional:dipart}
\begin{figure}[H]
    \centering
    \includegraphics[width=\linewidth]{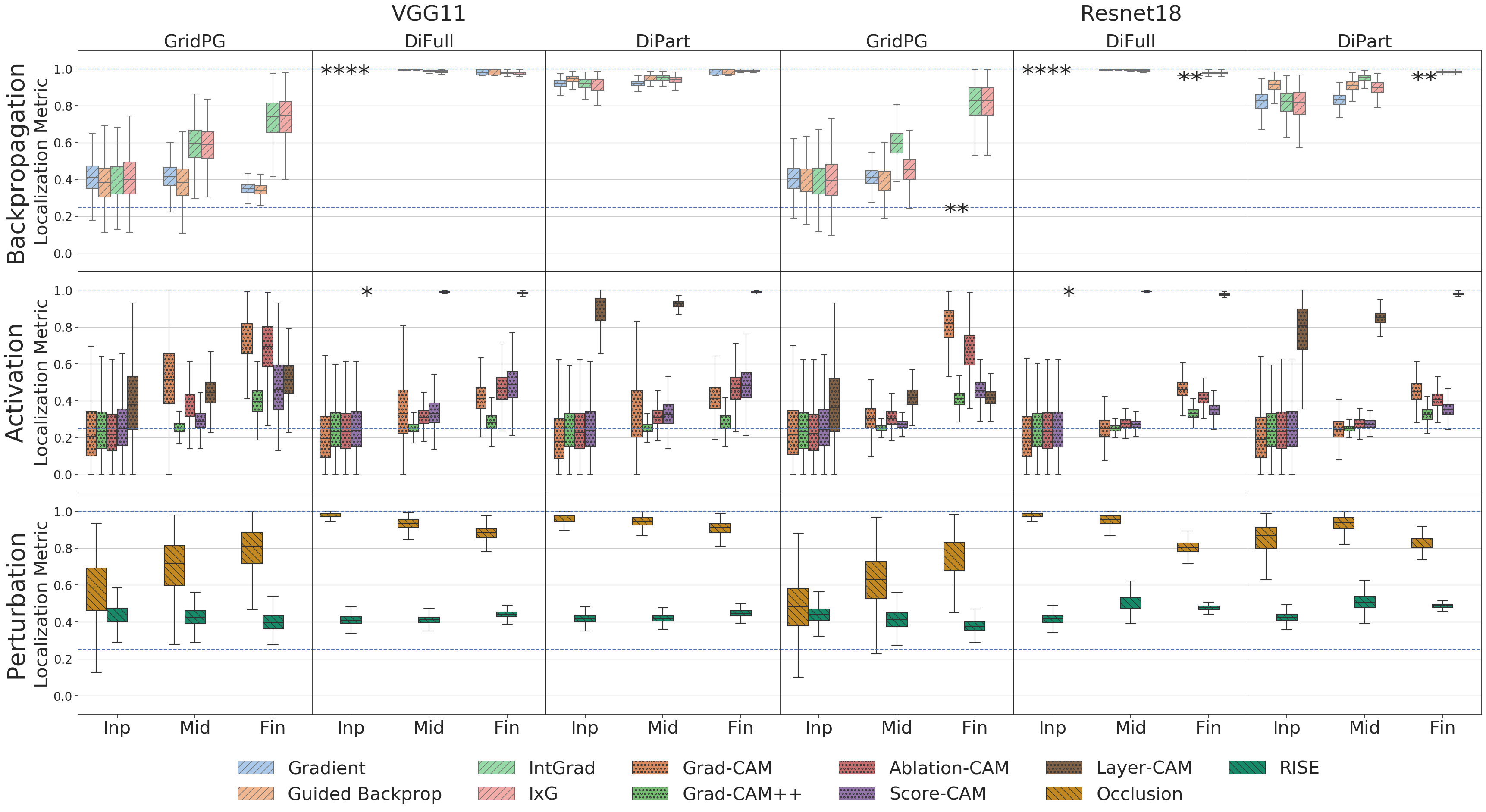}
    \caption{\textbf{Quantitative Results on ImageNet}. We evaluate the localization scores each attribution method at the input (Inp), middle (Mid), and final (Fin) convolutional layers, on each of \gridpg, \difull, and \dipart using VGG11 (left) and Resnet18 (right). \emph{Top:} Backpropagation-based methods. \emph{Middle:} Activation-based methods. \emph{Bottom:} Perturbation-based methods. The two horizontal dotted lines mark localization scores of $1.0$ and $0.25$, which correspond to perfect and random localization, respectively. We use the ``*'' symbol to show boxes that collapse to a single point, for better readability.}
    \label{fig:imagenet:quantitative}
\end{figure}

We provide the full results of our quantitative evaluation on the Grid Pointing Game \cite{boehle2021convolutional} (\gridpg), \difull, and \dipart using the backpropagation-based (\cref{fig:imagenet:quantitative}, top), activation-based (\cref{fig:imagenet:quantitative}, middle), and perturbation-based (\cref{fig:imagenet:quantitative}, bottom) methods on VGG11 \cite{simonyan2014very} (\cref{fig:imagenet:quantitative}, left) and Resnet18 \cite{he2016deep} (\cref{fig:imagenet:quantitative}, right).

It can be seen that the performance on \difull and \dipart is very similar across all three evaluation settings and the three layers. The most significant difference between the two can be seen among the backpropagation-based methods and \layercam\cite{jiang2021layercam} (\cref{fig:imagenet:quantitative}, row 1, cols. 2-3,5-6). On \difull, these methods show near-perfect localization, since the gradients of the outputs from each classification head that are used to assign importance are zero with respect to weights and activations of all grid cells disconnected from that head. On the other hand, the receptive field of the convolutional layers can overlap adjacent grid cells in \dipart, and the gradients of the outputs from the classification heads can thus have non-zero values with respect to inputs and activations from these adjacent grid regions. This also results in decreasing localization scores when moving backwards from the classifier.

Furthermore, the localization scores for \grad\cite{simonyan2013deep} and \gb\cite{springenberg2014striving} are constant at the final layer for Resnet18 (\cref{fig:imagenet:quantitative}, row 1, cols. 4-6). This is because this layer is immediately followed by a global average pooling layer, due to which all activations at this layer get an equal share of the gradients.

\newpage
\section{Qualitative Results using \aggatt}\label{sec:additional:aggatt}
In this section, we present additional qualitative results using our \aggatt evaluation along with examples of attributions from each bin, for each of \gridpg \cite{boehle2021convolutional} (\cref{sec:qual:gridpg}), \difull (\cref{sec:qual:difull}), and \dipart (\cref{sec:qual:dipart}).
\subsection{\gridpg}\label{sec:qual:gridpg}
\begin{figure}[H]
    \centering
    \includegraphics[width=0.6\linewidth]{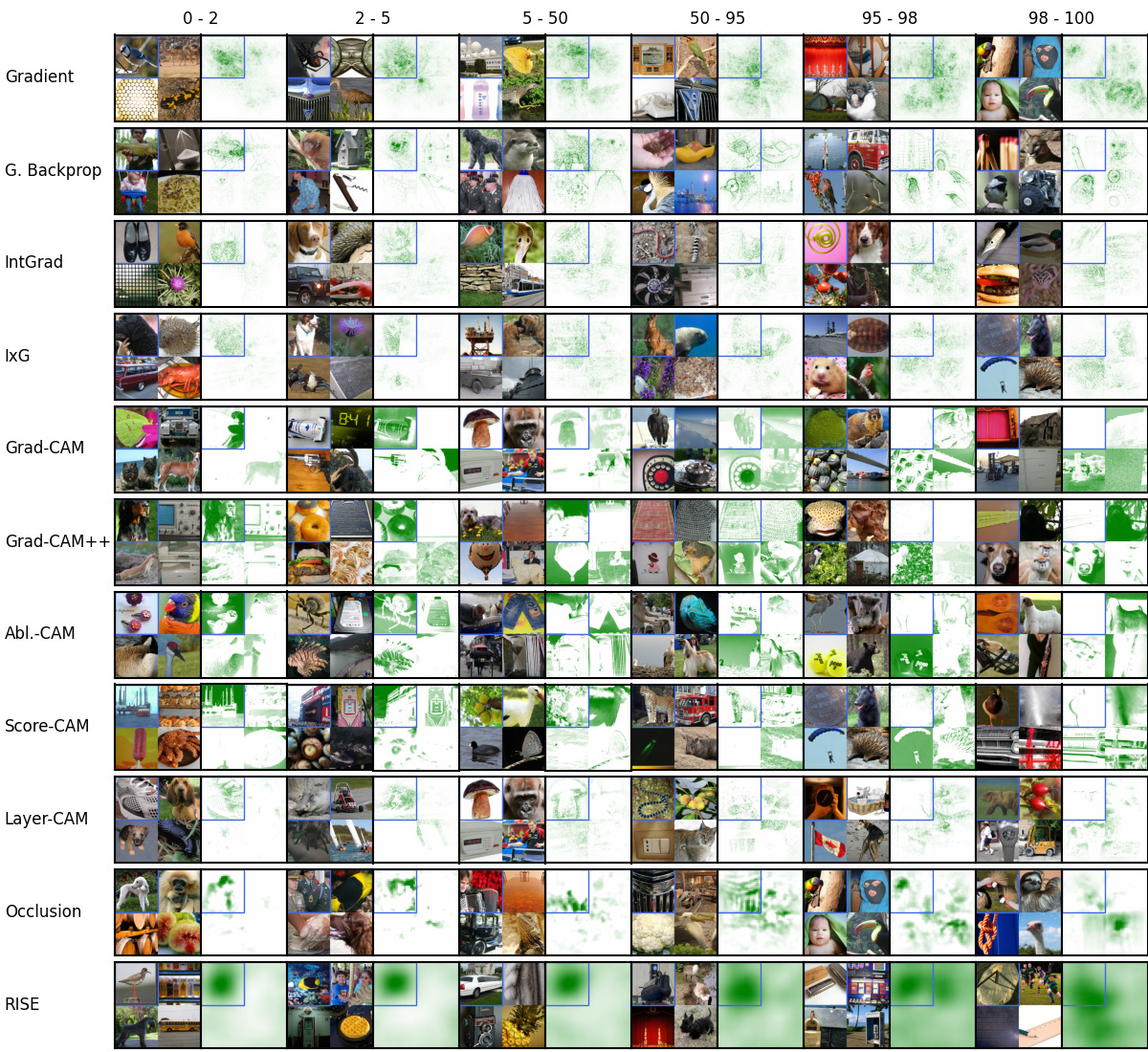}
    \caption{\textbf{Examples from each \aggatt bin for each method at the input layer on \gridpg using VGG11.} From each bin, the image and its attribution at the median position are shown.}
    \label{fig:vis:examples:gridpg:vgg11:input}
\end{figure}

\begin{figure}[H]
    \centering
    \includegraphics[width=0.6\linewidth]{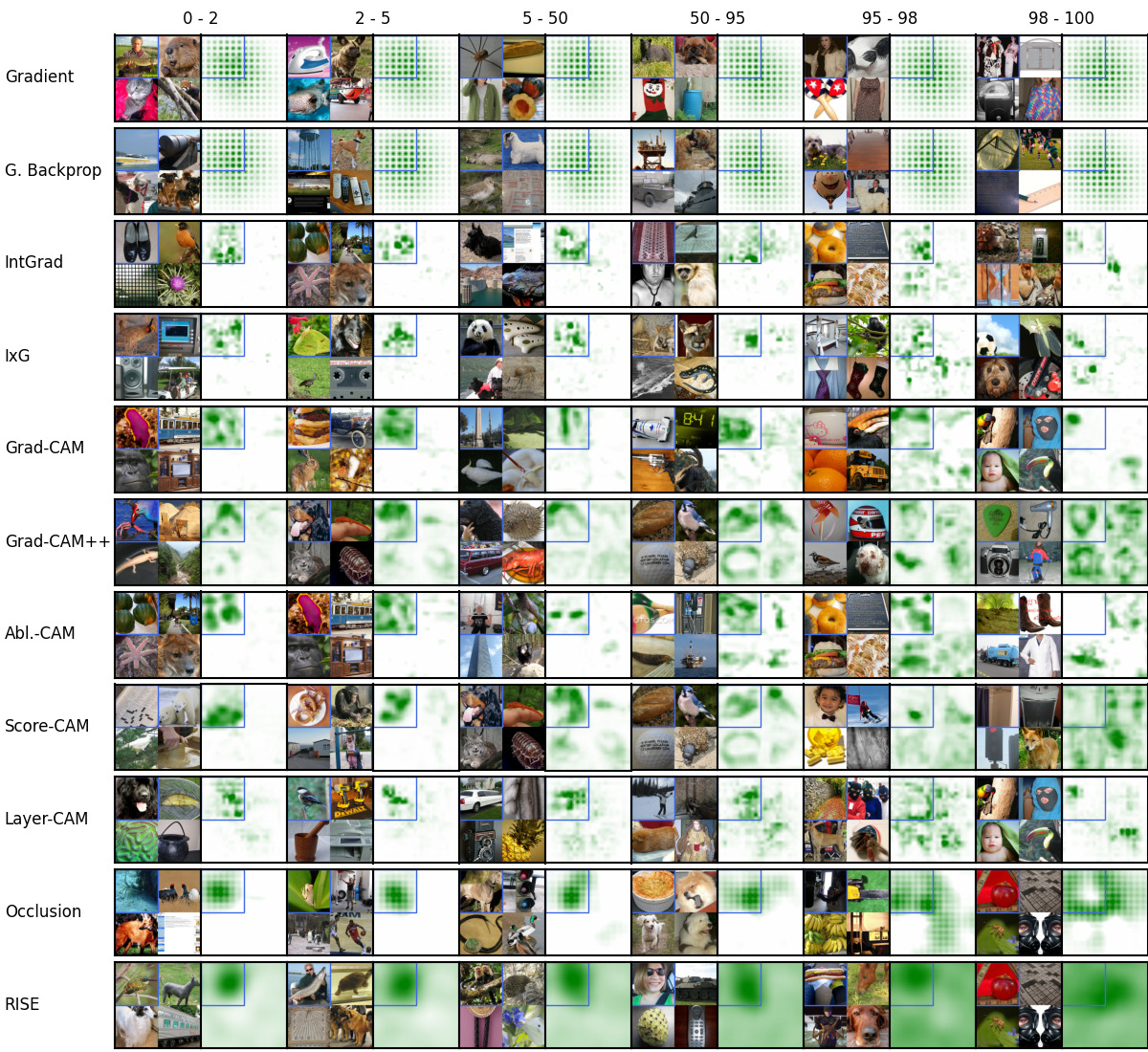}
    \caption{\textbf{Examples from each \aggatt bin for each method at the final layer on \gridpg using VGG11.} From each bin, the image and its attribution at the median position are shown.}
    \label{fig:vis:examples:gridpg:vgg11:final}
\end{figure}

\cref{fig:vis:examples:gridpg:vgg11:input} and \cref{fig:vis:examples:gridpg:vgg11:final} show examples from the median position of each \aggatt bin for each attribution method at the input and final layers, respectively, evaluated on \gridpg at the top-left grid cell using VGG11 \cite{simonyan2014very}. At the input layer (\cref{fig:vis:examples:gridpg:vgg11:input}), we observe that the \textbf{backpropagation-based methods} show noisy attributions that do not strongly localize to the top-left grid cell. This corroborates the poor quantitative performance of these methods at the input layer (\cref{fig:imagenet:quantitative}, top). With the exception of \layercam \cite{jiang2021layercam}, the \textbf{activation-based methods}, on the other hand, show strong attributions across all four grid cells, and localize very poorly. They appear to highlight the edges across the input irrespective of the class of each grid cell. This also agrees with the quantitative results (\cref{fig:imagenet:quantitative}, middle), where the median localization score of these methods is below the uniform attribution baseline. \layercam, being similar to \ixg \cite{shrikumar2017learning}, lies at the interface between activation and backpropagation-based methods, and also shows weak and noisy attributions. The \textbf{perturbation-based methods} visually show a high variance in attributions. While they localize well for about half the dataset (first three bins), the bottom half (last three bins) shows noisy and poorly localized attributions, which again agrees with the quantitative results (\cref{fig:imagenet:quantitative}, bottom). This further shows how evaluating on individual inputs can be misleading, and the utility of \aggatt for obtaining a holistic view across the dataset.

\begin{figure}[H]
    \centering
    \includegraphics[width=0.5\linewidth]{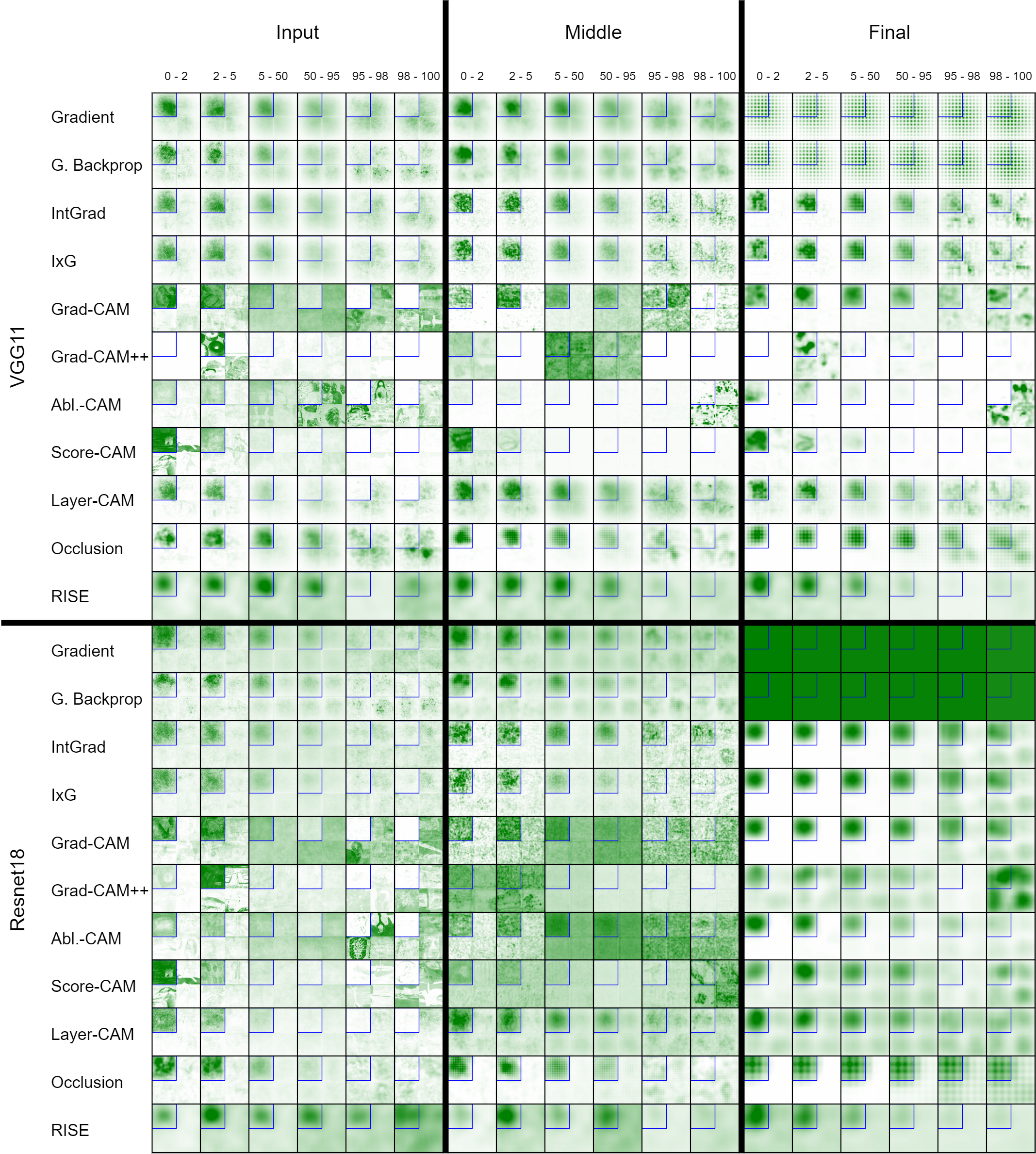}
    \caption{\textbf{\aggatt Evaluation on \gridpg for all methods at the input, middle, and final layers using VGG11 and Resnet18.}}
    \label{fig:vis:bins:gridpg}
\end{figure}

At the final layer (\cref{fig:vis:examples:gridpg:vgg11:final}), attributions from \grad \cite{simonyan2013deep} and \gb \cite{springenberg2014striving} are very noisy and only slightly concentrate at the top-left cell. The checkerboard-like pattern is a consequence of the max pooling operation after the final layer, which allocates all the gradient only to the maximum activation. Gradients from each position of the sliding classification kernel then get averaged to form the attributions. The localization of \intgrad \cite{sundararajan2017axiomatic}, \ixg, \gradcam \cite{selvaraju2017grad}, and \occ \cite{zeiler2013visualizing} improve considerably as compared to the input layer, which agrees with the quantitative results, and shows that diverse methods can show similar performance when compared fairly. The performance of the other activation-based methods and \rise \cite{petsiuk2018rise} improves to some extent, but is still poorly localized for around the half the dataset.
Finally, we show the \aggatt bins for all methods at all three layers using both VGG11 and Resnet18 \cite{he2016deep} in \cref{fig:vis:bins:gridpg}. We see that the \aggatt bins reflect the trends observed in the examples in each bin, and serve as a useful tool for visualization.

\subsection{\difull}\label{sec:qual:difull}
\begin{figure}[H]
    \centering
    \includegraphics[width=0.6\linewidth]{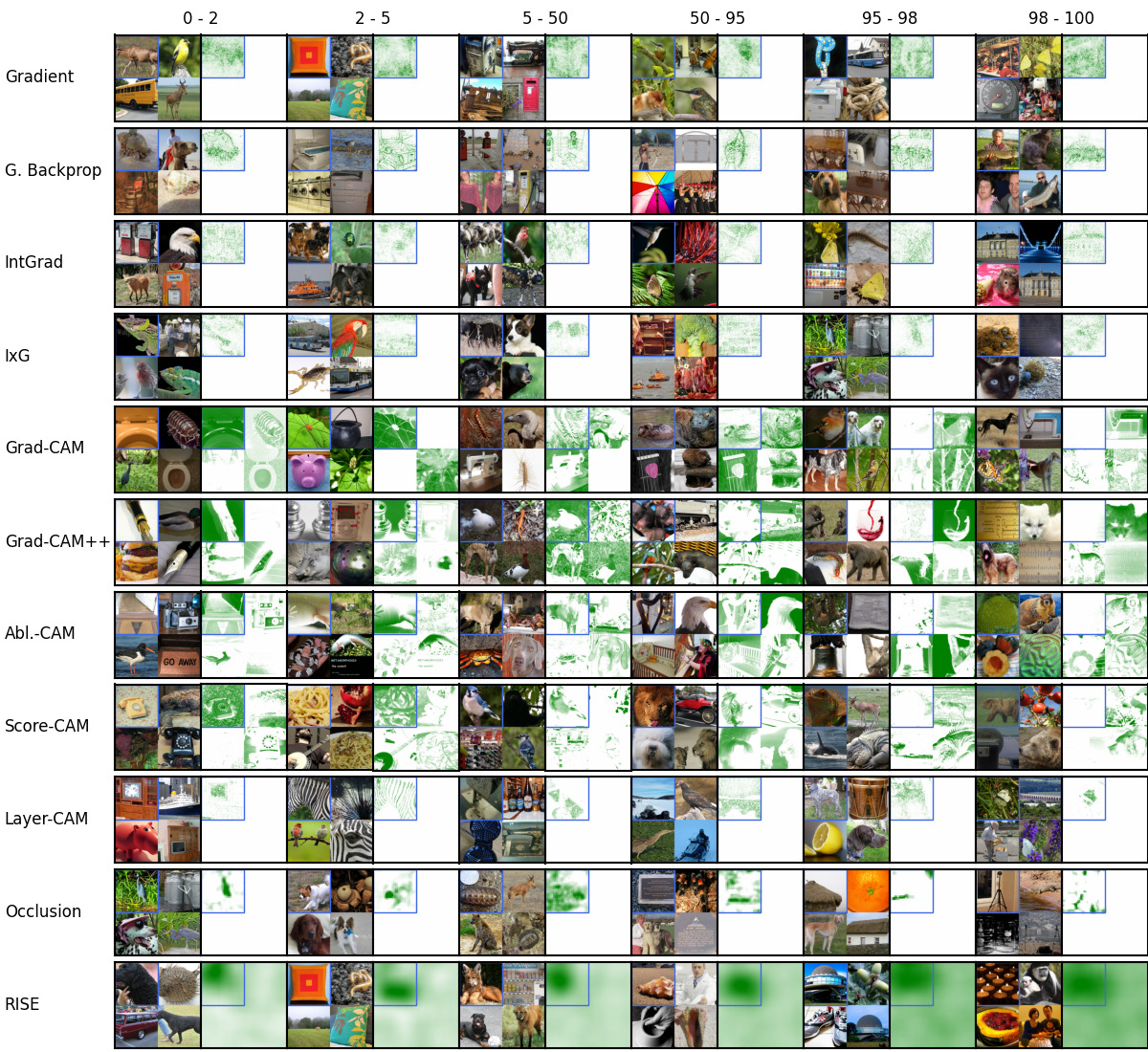}
    \caption{\textbf{Examples from each \aggatt bin for each method at the input layer on \difull using VGG11.} From each bin, the image and its attribution at the median position are shown.}
    \label{fig:vis:examples:difull:vgg11:input}
\end{figure}

\begin{figure}[H]
    \centering
    \includegraphics[width=0.6\linewidth]{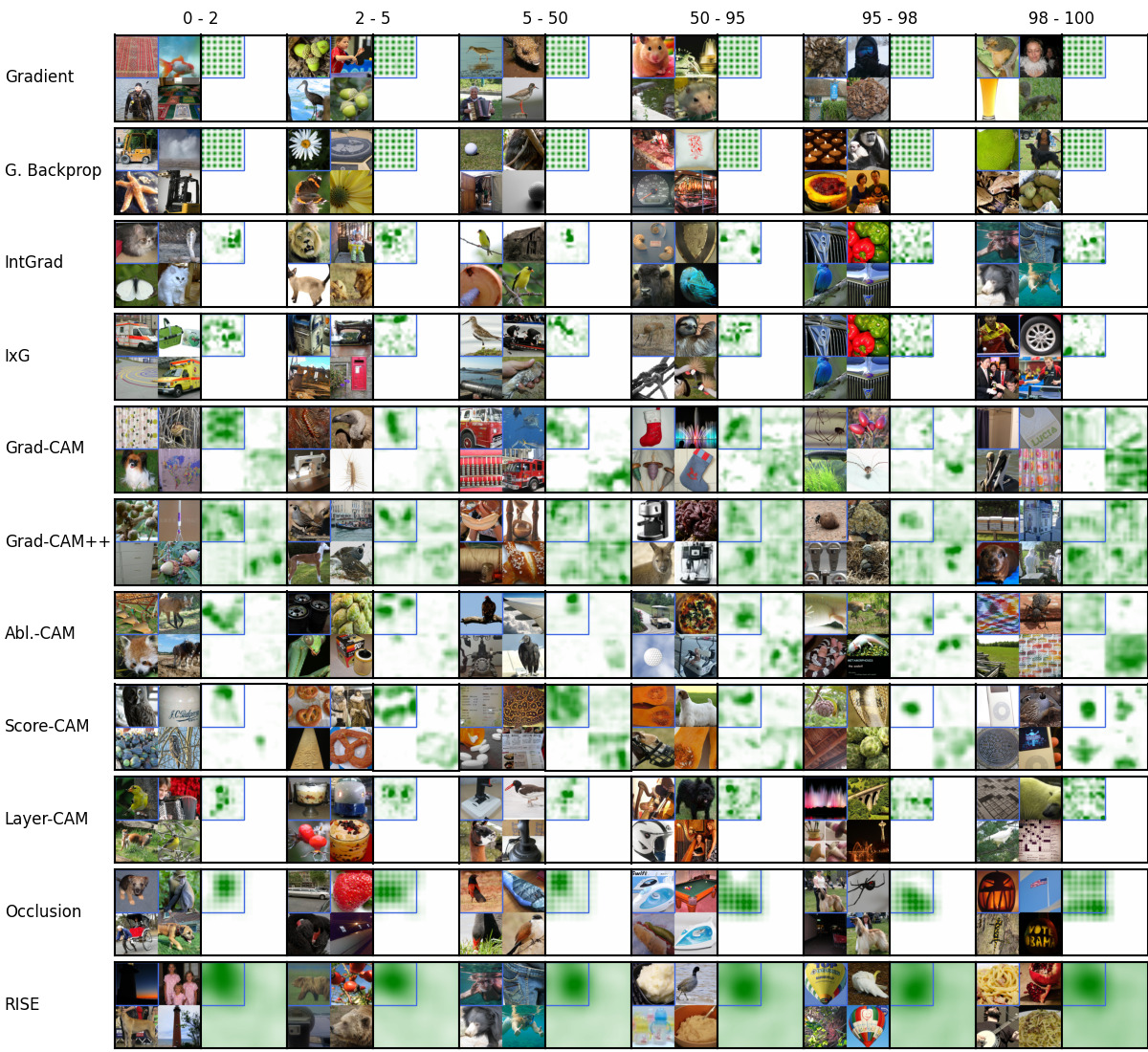}
    \caption{\textbf{Examples from each \aggatt bin for each method at the final layer on \difull using VGG11.} From each bin, the image and its attribution at the median position are shown.}
    \label{fig:vis:examples:difull:vgg11:final}
\end{figure}

\cref{fig:vis:examples:difull:vgg11:input} and \cref{fig:vis:examples:difull:vgg11:final} show examples from the median position of each \aggatt bin for each attribution method at the input and final layers, respectively, evaluated on \difull at the top-left grid cell using VGG11. At the input layer (\cref{fig:vis:examples:difull:vgg11:input}), the backpropagation-based methods and \layercam show perfect localization across the dataset. This is explained by the disconnected construction of \difull, and agrees with the quantitative results shown in \cref{fig:imagenet:quantitative}). The activation-based methods show very poor localization that appear visually similar to the attributions observed on \gridpg (\cref{sec:qual:gridpg}). \occ shows near-perfect localization, since the placement of the classification kernel at any location not overlapping with the top-left grid cell does not influence the output in the \difull setting. \rise still produces noisy attributions across the dataset. While only the top-left grid cell influences the output, the use of random masks causes input regions that share masks with inputs in the top-left cell to also get attributed.

\begin{figure}[H]
    \centering
    \includegraphics[width=0.5\linewidth]{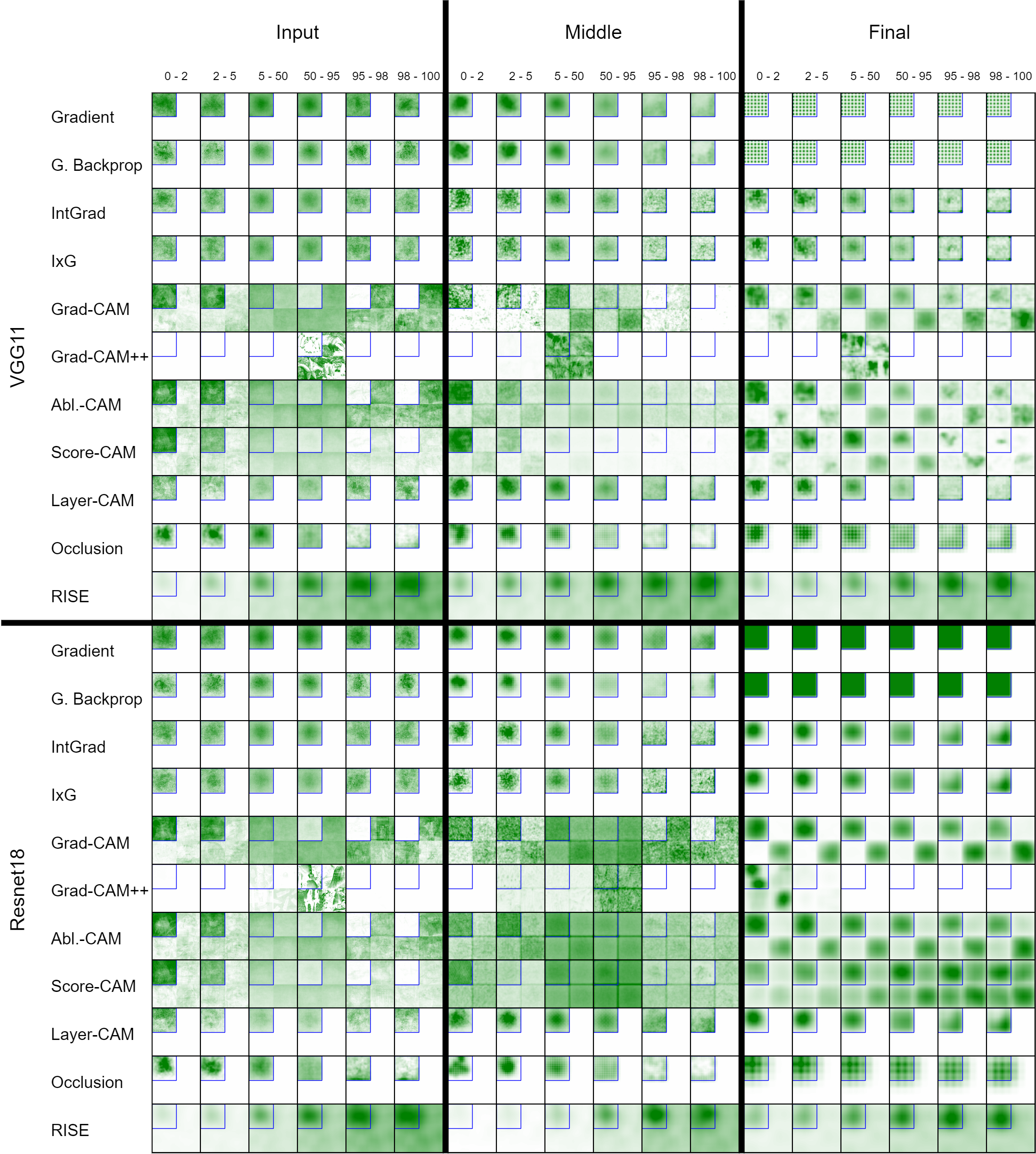}
    \caption{\textbf{\aggatt Evaluation on \difull for all methods at the input, middle, and final layers using VGG11 and Resnet18.}}
    \label{fig:vis:bins:difull}
\end{figure}

At the final layer (\cref{fig:vis:examples:difull:vgg11:final}), the backpropagation-based methods and \layercam still show perfect localization, for the same reason as discussed above. Attributions from \grad and \gb show similar artifacts as seen with \gridpg (\cref{sec:qual:gridpg}), but are localized to the top-left cell. The activation-based methods apart from \layercam concentrate their attributions at the top-left and bottom-right grid cells, particularly in the early bins. This is because both these cells contain images from the same class, and the weighing of activation maps by these methods using a scalar value causes both to be attributed, even though only the instance at the top-left influences the classification. Further, \occ and \rise show similar results as at the input layer. The attributions of \occ are noticeably lower in resolution, since the relative size of the occlusion kernel as compared to the activation map is much larger at the final layer.
Finally, we show the \aggatt bins for all methods at all three layers using both VGG11 and Resnet18 in \cref{fig:vis:bins:difull}, and see that they reflect the trends observed in the individual examples seen from each bin.

\subsection{\dipart}\label{sec:qual:dipart}
\begin{figure}[H]
    \centering
    \includegraphics[width=0.6\linewidth]{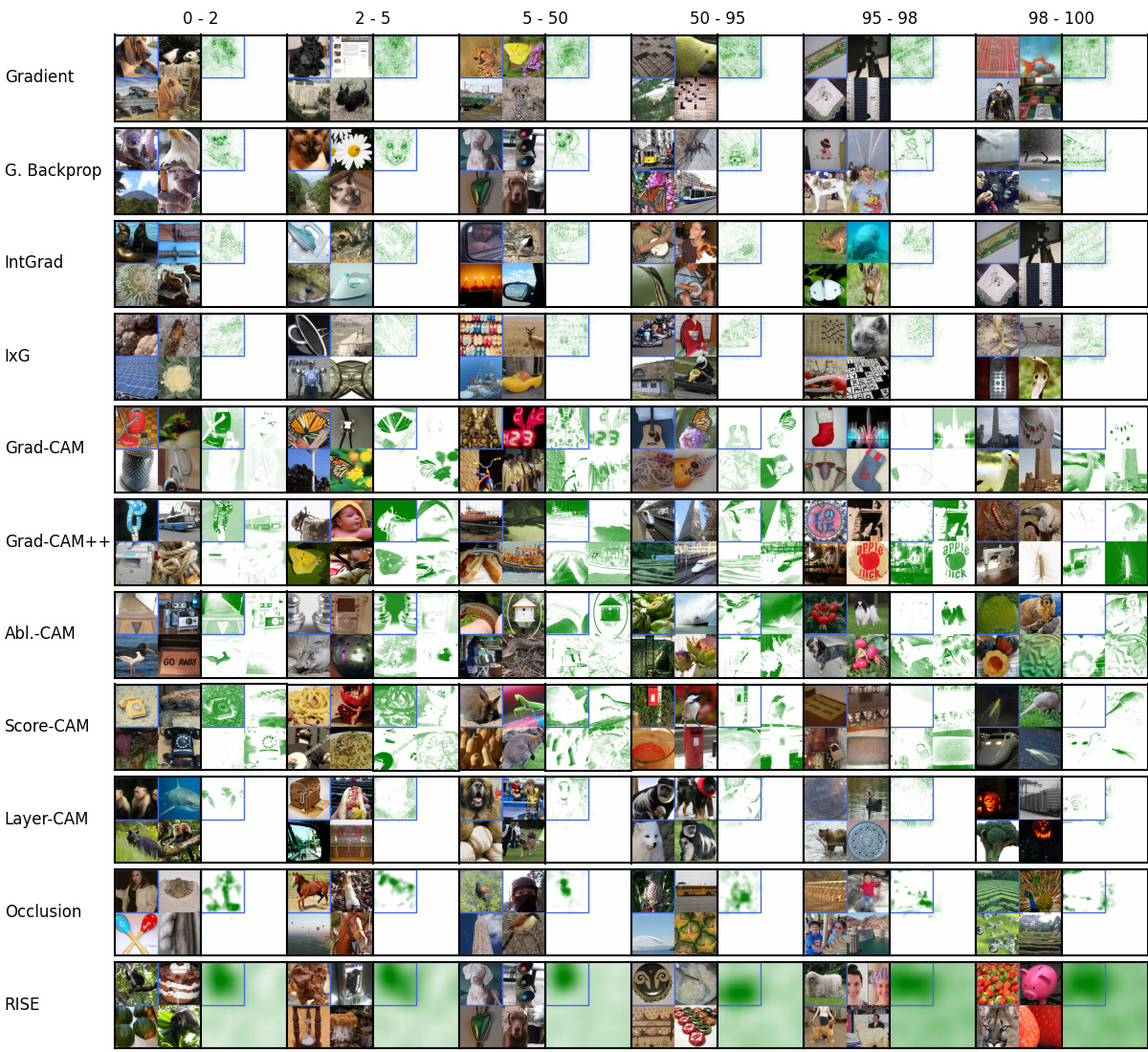}
    \caption{\textbf{Examples from each \aggatt bin for each method at the input layer on \dipart using VGG11.} From each bin, the image and its attribution at the median position are shown.}
    \label{fig:vis:examples:dipart:vgg11:input}
\end{figure}

\begin{figure}[H]
    \centering
    \includegraphics[width=0.6\linewidth]{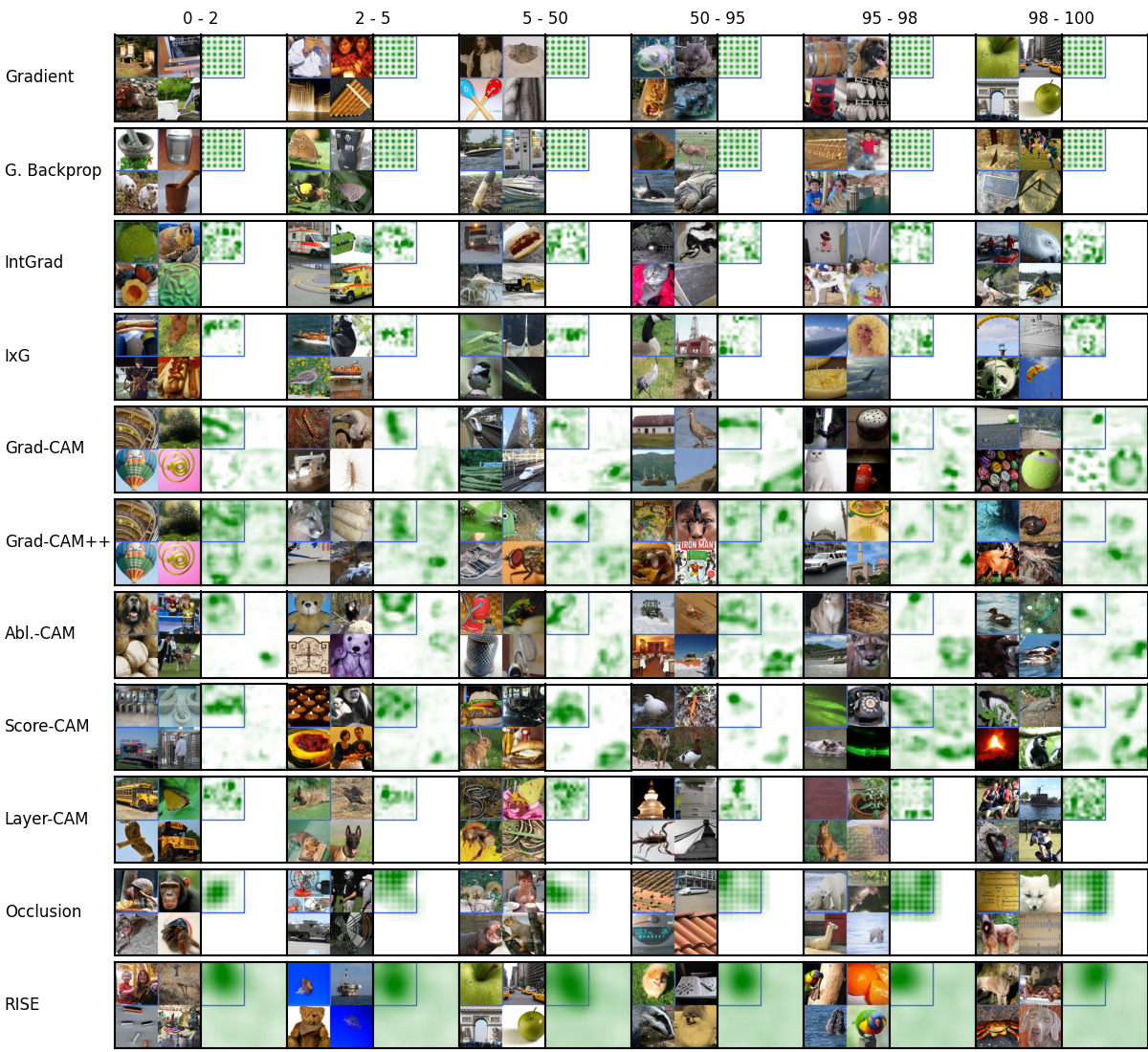}
    \caption{\textbf{Examples from each \aggatt bin for each method at the final layer on \dipart using VGG11.} From each bin, the image and its attribution at the median position are shown.}
    \label{fig:vis:examples:dipart:vgg11:final}
\end{figure}

\begin{figure}[H]
    \centering
    \includegraphics[width=0.5\linewidth]{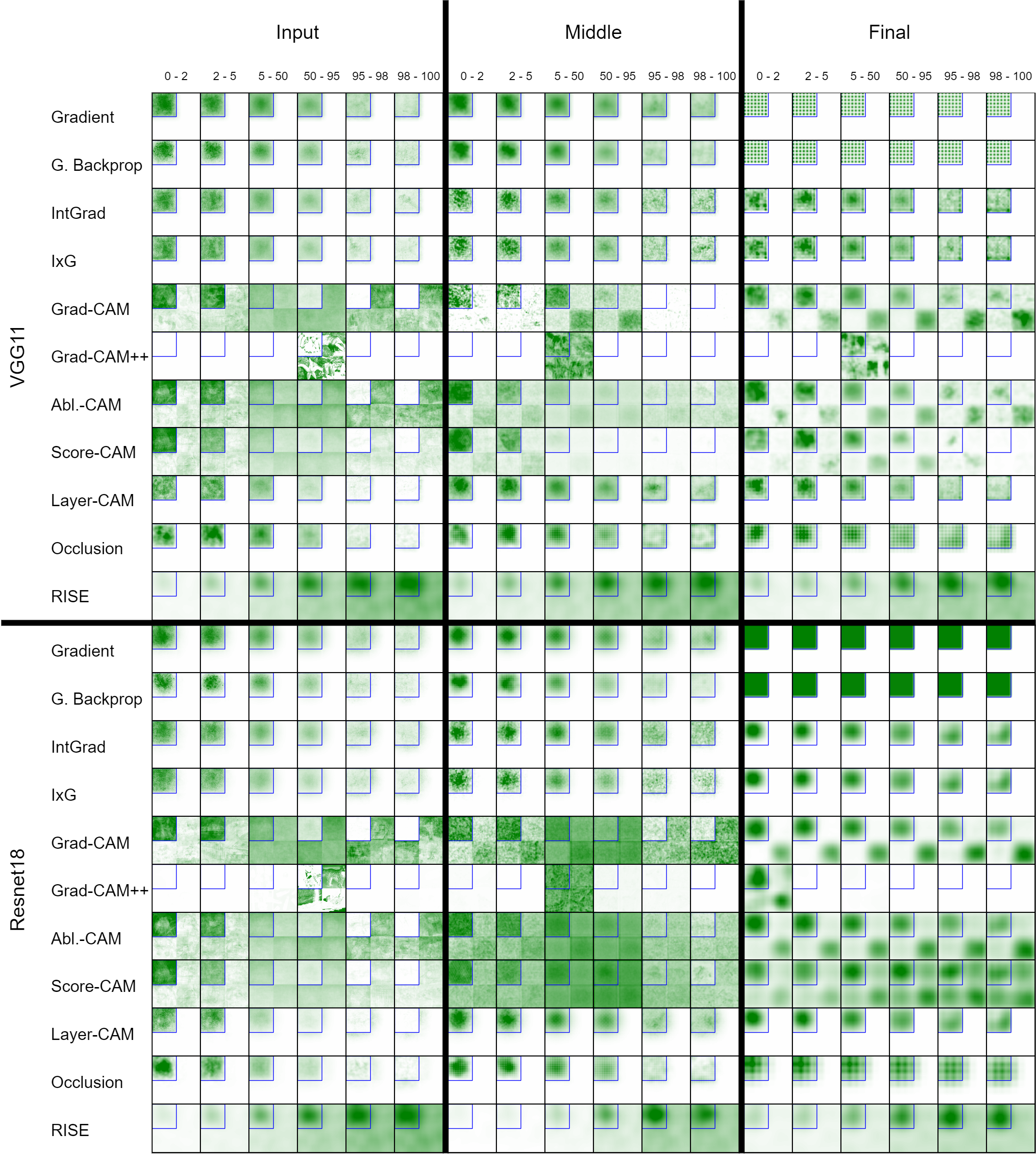}
    \caption{\textbf{\aggatt Evaluation on \dipart for all methods at the input, middle, and final layers using VGG11 and Resnet18.}}
    \label{fig:vis:bins:dipart}
\end{figure}

\cref{fig:vis:examples:dipart:vgg11:input} and \cref{fig:vis:examples:dipart:vgg11:final} show examples from the median position of each \aggatt bin for each attribution method at the input and final layers, respectively, evaluated on \dipart at the top-left grid cell using VGG11. In addition, \cref{fig:vis:bins:dipart} shows the \aggatt bins for all methods at all three layers using both VGG11 and Resnet18. As observed with the quantitative results (\cref{sec:additional:dipart}, the performance seen visually on \dipart across the three layers is very similar to that on \difull (\cref{sec:qual:difull}). However, they slightly differ in the case of the backpropagation-based methods and \layercam, particularly at the input layer (\cref{fig:vis:examples:dipart:vgg11:input}). This is because unlike in \difull, the grid cells are only partially disconnected, and the receptive field of the convolutional layers can overlap adjacent grid cells to some extent. Nevertheless, as can be seen here, only a small boundary region around the top-left grid cell receives attributions, and the difference is not visually very perceivable. This further shows that the \dipart setting can be thought of as a natural extension for \difull, that mostly shares the requisite property without being an entirely constructed setting.

\newpage
\section{Correlation between Attributions}\label{sec:additional:correlation}
\begin{figure}[H]
    \centering
    \includegraphics[width=\linewidth]{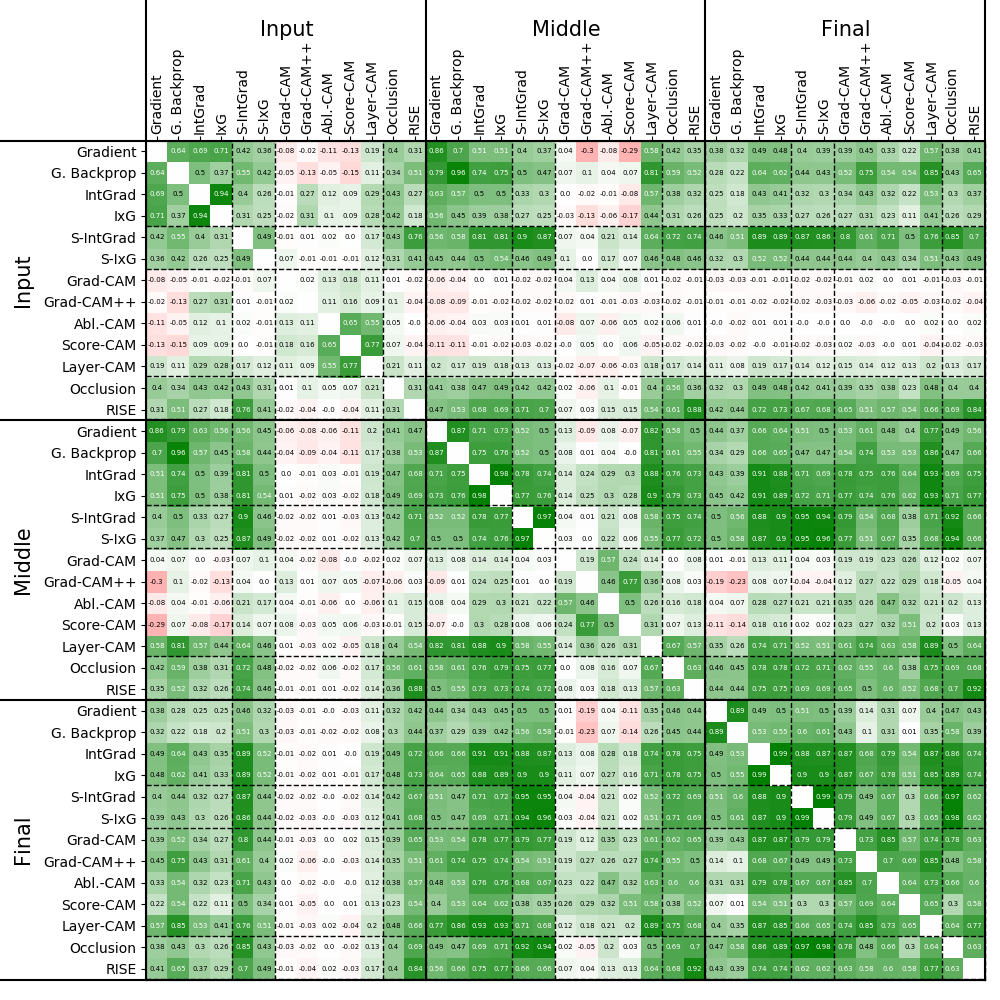}
    \caption{\textbf{Spearman rank correlation coefficients between each pair of methods at all three layers on \gridpg for VGG11.} We observe that diverse methods (\intgrad, \ixg, \gradcam, \occ) correlate strongly when evaluated at the final layer (bottom-right), which agrees with their similar quantitative and \aggatt performance. Further, \sintgrad and \sixg at the input layer correlate significantly stronger with \gradcam at the final layer as compared to \intgrad and \ixg at the input layers. This, combined with the similarity in quantitative and \aggatt performance, shows the utility of smoothing in obtaining attributions that localize well while also performing much better in not attributing impossible regions specified by \difull.}
    \label{fig:correlation}
\end{figure}

From the quantitative (\cref{fig:imagenet:quantitative}) and qualitative (\cref{fig:vis:bins:gridpg}) results, we observed that diverse methods perform similarly on \gridpg \cite{boehle2021convolutional} both in terms of localization score and through \aggatt visualizations when evaluated fairly. This was particularly the case with \intgrad \cite{sundararajan2017axiomatic}, \ixg \cite{shrikumar2017learning}, \gradcam \cite{selvaraju2017grad}, and \occ \cite{zeiler2013visualizing}, when evaluated at the final layer. We also found (Sec 5.2 in the paper) that smoothing \intgrad and \ixg (the result of which we call \sintgrad and \sixg) attributions evaluated at the \emph{input layer} leads to visually and quantitatively similar performance as \gradcam evaluated at the \emph{final layer}. In this section, we investigate this further, and study the correlation of these methods at the level of individual attributions. In particular, we compute the Spearman rank correlation coefficient of the localization scores using VGG11 \cite{simonyan2014very} of every pair of methods from each of the three layers. The results are shown in \cref{fig:correlation}.

We observe that at the input layer (\cref{fig:correlation}, top-left corner), the activation-based methods are poorly correlated with each other and with the backpropagation and perturbation-based methods. This also agrees with the poor localization of these methods seen previously (\cref{fig:imagenet:quantitative}, \cref{fig:vis:examples:gridpg:vgg11:input}). The backpropagation-based and perturbation-based methods, on the other hand, show moderate to strong correlation amongst and with each other. Similar results can be seen when comparing methods at the middle layer with the input layer and the final layer (\cref{fig:correlation}, edge centres).  However, when compared at the middle layer (\cref{fig:correlation}, middle), the activation-based methods still correlate poorly with other methods, but the strength of the correlation improves in general.

Further, when compared at the final layer (\cref{fig:correlation}, bottom-right corner), all methods show moderate to strong correlations with each other. This could be because generating explanations at the final layer is a significantly easier task as compared to doing so at the input, since the activations are used as is and only the classification layers' outputs are explained. The pairs with very strong positive correlation also show that attribution methods with diverse mechanisms can perform similarly when evaluated fairly. Finally, we observe that the activation-based methods at the final layer, instead of the input layer, correlate much better with the other methods at the input layer (\cref{fig:correlation}, top-right, bottom-left).

We also observe that \sintgrad and \sixg at the input layer correlate well with the best-performing methods (\intgrad, \ixg, \gradcam, \occ) at the final layer. Further, this marks a significant improvement when compared with \intgrad and \ixg at the input layer. For example, \intgrad at the input layer compared with \gradcam at the final layer results in a correlation coefficient of $0.34$, while \sintgrad results in a correlation coefficient of $0.80$.

We further study the effect of smoothing in \cref{tab:corr:smooth:IxG,tab:corr:smooth:IntGrad}. We observe that the correlation between \sintgrad and \sixg improves significantly over \intgrad and \ixg for VGG11 when using large kernels. However, for Resnet18 \cite{he2016deep}, the improvement for \sixg is very small. This agrees with the quantitative localization performance of these methods (Sec 5.2 in the paper). This shows that beyond aggregate visual similarity and quantitative performance, smoothing \intgrad and \ixg can produce explanations at the input layer that are individually similar to \gradcam at the final layer, while also explaining the full network and performing significantly better on \difull. We further visually compare the impact of smoothing in \cref{sec:additional:smoothing}.

\begin{figure}[H]
\centering
\begin{tabular}{|l|c|c|c|c|c|c|c|}
\hline
 & Original & $K=9$ & $K=17$ & $K=33$ & $K=65$ & $K=129$ & $K=257$ \\
\hline
\hline
VGG11 & 0.34 & 0.42 & 0.52 & 0.69 & 0.78 & \textbf{0.80} & 0.71 \\
\hline
Resnet18 & 0.18 & 0.21 & 0.27 & 0.40 & 0.55 & \textbf{0.63} & 0.61 \\
\hline
\end{tabular}
\captionof{table}{\textbf{Spearman rank correlation coefficients between \gradcam at the final layer and S-IntGrad at the input layer on \gridpg for varying degrees of smoothing.} The first column shows the correlation with the original unsmoothed version. We observe that the correlation improves significantly for both VGG11 and Resnet18 when smoothing with large kernel sizes.}
\label{tab:corr:smooth:IntGrad}
\end{figure}
\begin{figure}[H]
\centering
\begin{tabular}{|l|c|c|c|c|c|c|c|}
\hline
& Original & $K=9$ & $K=17$ & $K=33$ & $K=65$ & $K=129$ & $K=257$ \\
\hline
\hline
VGG11 & 0.27 & 0.28 & 0.33 & 0.43 & \textbf{0.49} & 0.44 & 0.34 \\
\hline
Resnet18 & 0.14 & 0.13 & 0.15 & 0.17 & \textbf{0.18} & 0.13 & 0.05 \\
\hline
\end{tabular}
\captionof{table}{\textbf{Spearman rank correlation coefficients between \gradcam at the final layer and S-IxG at the input layer on \gridpg for varying degrees of smoothing.} The first column shows the correlation with the original unsmoothed version. We observe that the correlation improves for VGG11, but does not significantly improve for Resnet18.}
\label{tab:corr:smooth:IxG}
\end{figure}

\newpage
\section{Impact of Smoothing Attributions}\label{sec:additional:smoothing}
In this section, we explore the impact of smoothing attributions. First, we briefly discuss a possible reason for the improvement in localization of attributions after smoothing (\cref{sec:smooth:explain}. Then, we visualize the impact of smoothing through examples and \aggatt visualizations (\cref{sec:smooth:aggatt}). Further, we also compare the performance of \gradcam \cite{selvaraju2017grad} at the final layer with \sintgrad and \sixg at the input layer across the same examples from each bin and show their similarities across bins (\cref{sec:smooth:compare}).

\subsection{Effect of Smoothing}\label{sec:smooth:explain}
We believe that our smoothing results highlight an interesting aspect of piece-wise linear models (PLMs), which goes beyond mere practical improvements. For PLMs (such as the models used here), \ixg \cite{shrikumar2017learning} yields the exact pixel contributions according to the linear mapping given by the PLM. In other words, the sum of \ixg attributions over all pixels yields exactly (ignoring biases) the model output. If the effective receptive field of the model is small (cf. \cite{luo2016understanding}), sum pooling \ixg with a kernel of the same size accurately computes the model’s local output (apart from the influence of bias terms). Our method of smoothing \ixg with a Gaussian kernel performs a weighted average pooling of attributions in the local region around each pixel, which produces a similar effect and appears to summarize the effect of the pixels in the local region to the model's output, which leads to less noisy attributions and better localization.

\subsection{\aggatt Evaluation after Smoothing}\label{sec:smooth:aggatt}
\begin{figure}[H]
	\centering
		\includegraphics[width=0.9\linewidth]{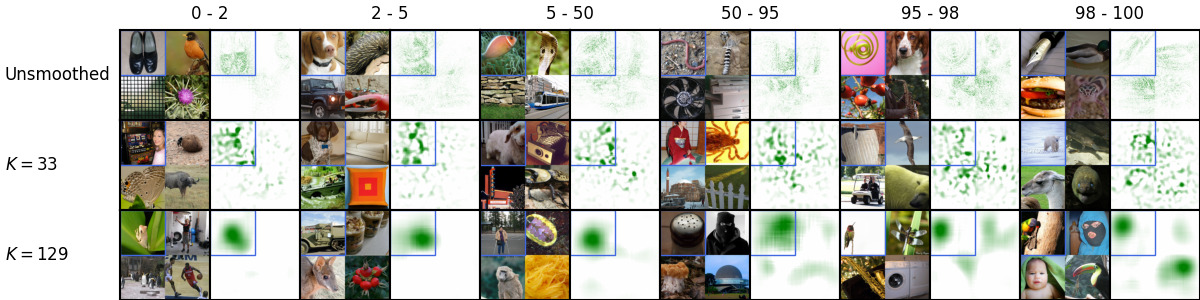}
		\caption{\textbf{Examples from each \aggatt bin after smoothing \intgrad attributions on \gridpg using VGG11.} From each bin, the image and its attribution at the median position are shown.}
		\label{fig:vis:examples:smooth:gridpg:vgg11:intgrad}
\end{figure}

\begin{figure}[H]
		\centering
		\includegraphics[width=0.9\linewidth]{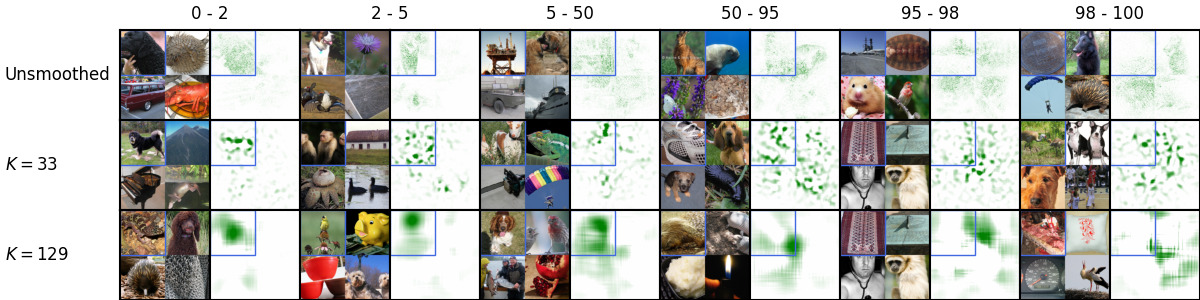}
		\caption{\textbf{Examples from each \aggatt bin after smoothing \ixg attributions on \gridpg using VGG11.} From each bin, the image and its attribution at the median position are shown.}
		\label{fig:vis:examples:smooth:gridpg:vgg11:ixg}
\end{figure}

\begin{figure}[H]
	\centering
	\includegraphics[width=0.5\linewidth]{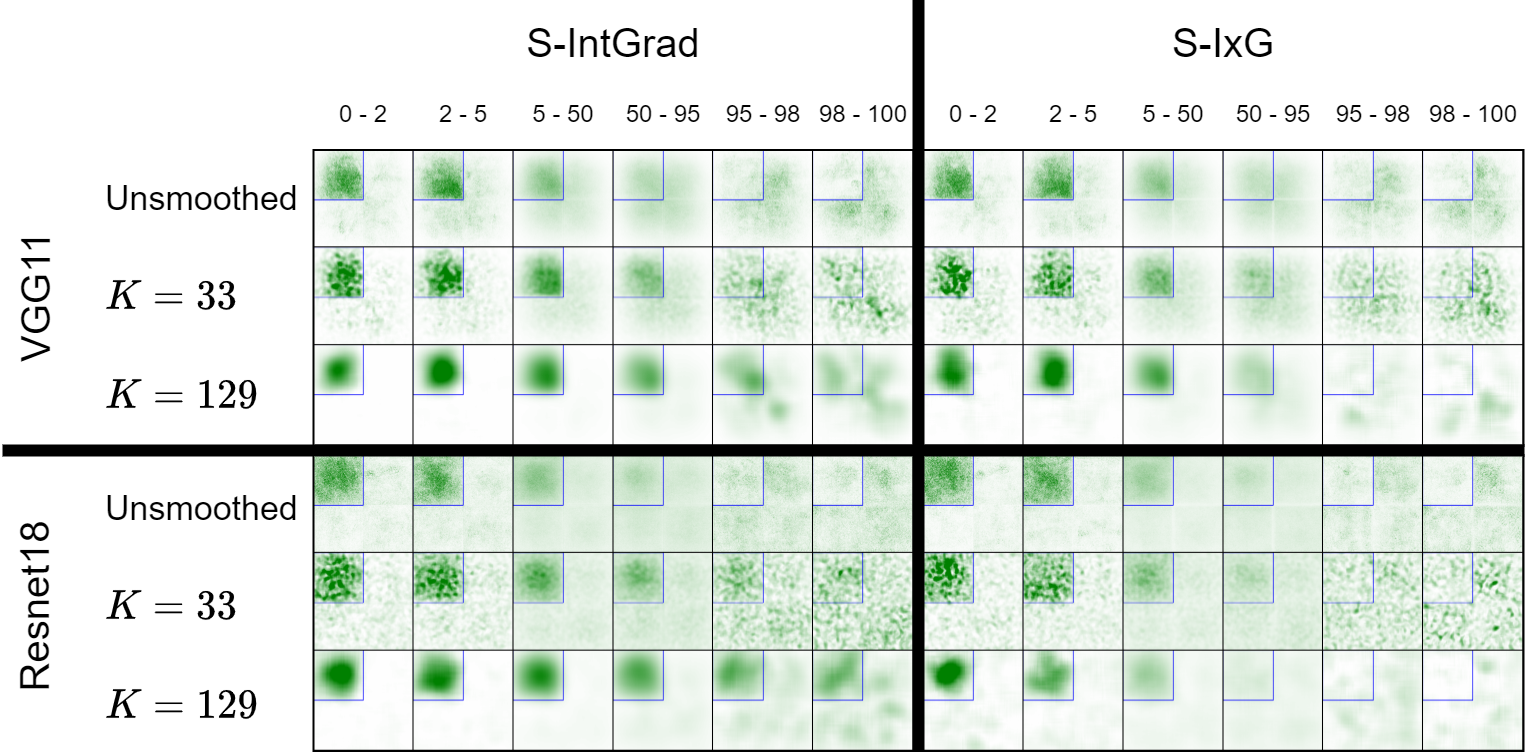}
	\caption{\textbf{Impact of smoothing \intgrad and \ixg attributions using VGG11 and Resnet18 visualized through the \aggatt evaluation on \gridpg.}}
	\label{fig:vis:bins:smooth:gridpg}
\end{figure}

In \cref{fig:vis:examples:smooth:gridpg:vgg11:intgrad} and \cref{fig:vis:examples:smooth:gridpg:vgg11:ixg}, we show examples from each \aggatt bin for \sintgrad and \sixg at the input layer for two different kernel sizes, and compare with \intgrad \cite{sundararajan2017axiomatic} and \ixg at the input layer respectively. We observe that the localization performance significantly improves with increasing kernel size, and produces much stronger attributions for the target grid cell. In \cref{fig:vis:bins:smooth:gridpg}, we show the \aggatt bins for these methods on both VGG11 \cite{simonyan2014very} and Resnet18 \cite{he2016deep}. We see that this reflects the trends seen from the examples, and also clearly shows the relative ineffectiveness of smoothing \ixg for Resnet18 (\cref{fig:vis:bins:smooth:gridpg} bottom right and \cref{tab:corr:smooth:IxG}).
\subsection{Comparing \gradcam with \sintgrad and \sixg}\label{sec:smooth:compare}
\begin{figure}[H]
	\centering
	\begin{subfigure}{0.49\linewidth}
	\centering
	    \includegraphics[width=\linewidth]{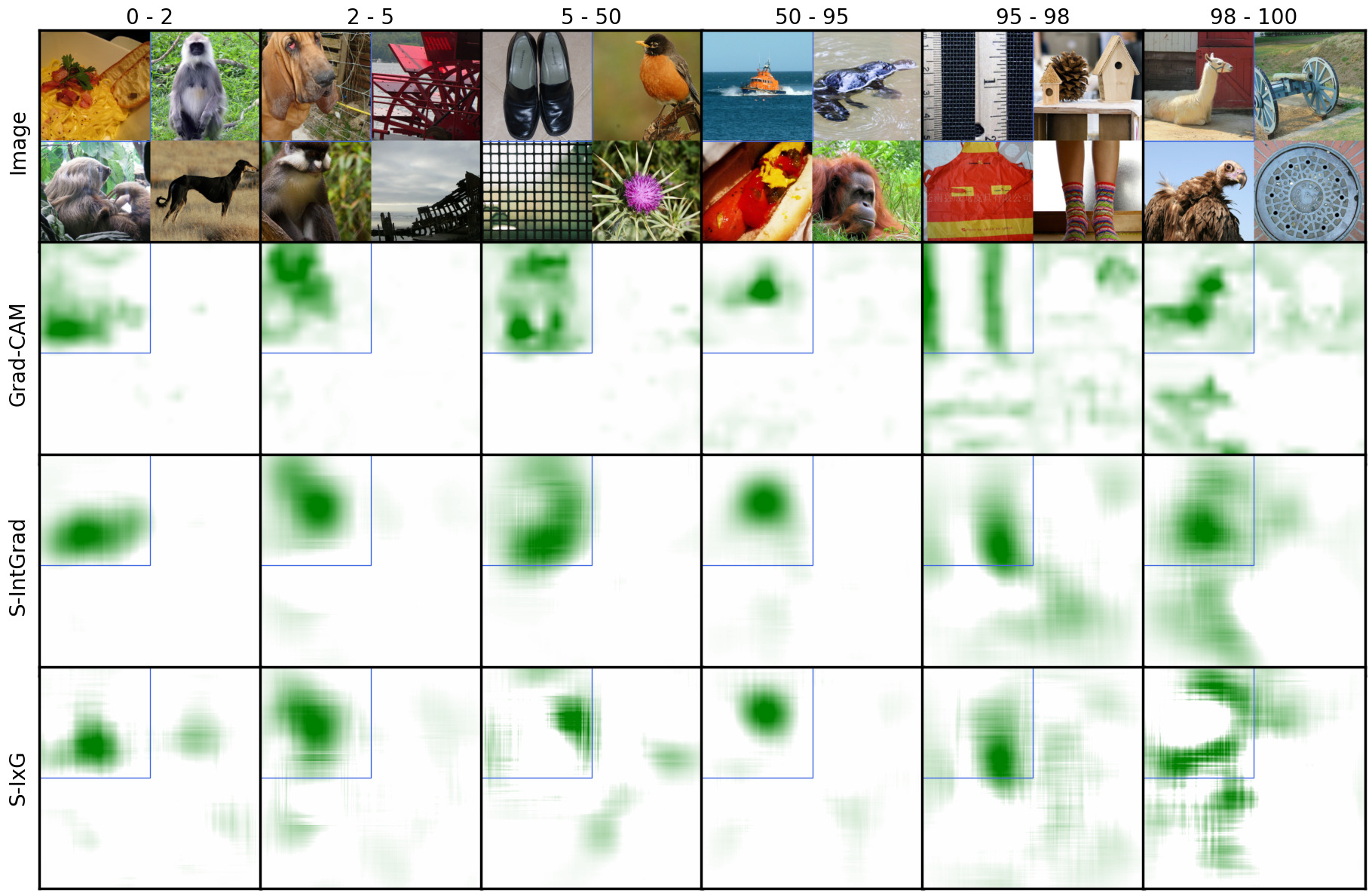}
	\caption{VGG11}
	\label{fig:vis:examples:smooth:compare:vgg11}
	\end{subfigure}
	\hfill
	\begin{subfigure}{0.49\linewidth}
	\centering
	\includegraphics[width=\linewidth]{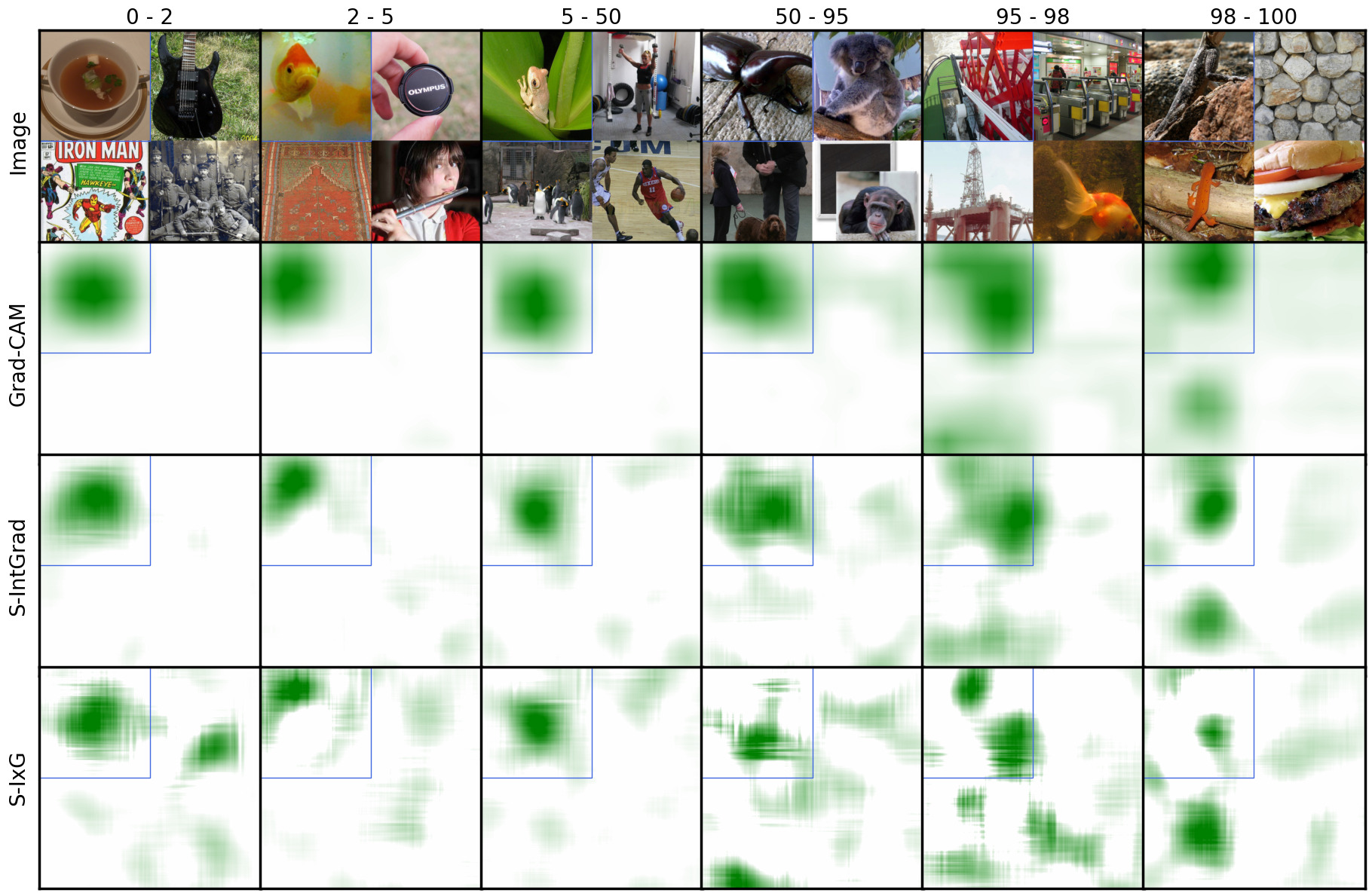}
	\caption{Resnet18}
	\label{fig:vis:examples:smooth:compare:resnet18}
	\end{subfigure}
	\caption{\textbf{Example attributions from each \aggatt bin of \gradcam at the final layer compared with corresponding attributions from \sintgrad and \sixg at the input layer with $K=129$, using Resnet18 on \gridpg.} We observe that \sintgrad and \sixg show visually similar examples to \gradcam across bins for VGG11. While \sintgrad also performs similarly for Resnet18, \sixg produces more noisy attributions.}
	\label{fig:vis:examples:smooth:compare}
\end{figure}

We now compare \gradcam at the final layer with \sintgrad and \sixg at the input layer with $K=129$ on the same set of examples (\cref{fig:vis:examples:smooth:compare}). We pick an example from each \aggatt bin of \gradcam, and evaluate all three methods on them. From \cref{fig:vis:examples:smooth:compare}, we observe that the three methods produce visually similar attributions across the \aggatt bins. While the attributions of \sintgrad and \sixg are somewhat coarser than \gradcam, particularly for the examples in the first few bins, they still concentrate around similar regions in the images. Interestingly, they perform similarly even for examples where \gradcam does not localize well, \ie, in the last two bins. Finally, we again see that \sixg using Resnet18 performs relatively worse as compared to the other methods (as also seen in \cref{tab:corr:smooth:IxG}).

\newpage
\section{Quantitative Evaluation on All Layers}\label{sec:alllayers}
\begin{figure}[H]
    \centering
    \includegraphics[width=\linewidth]{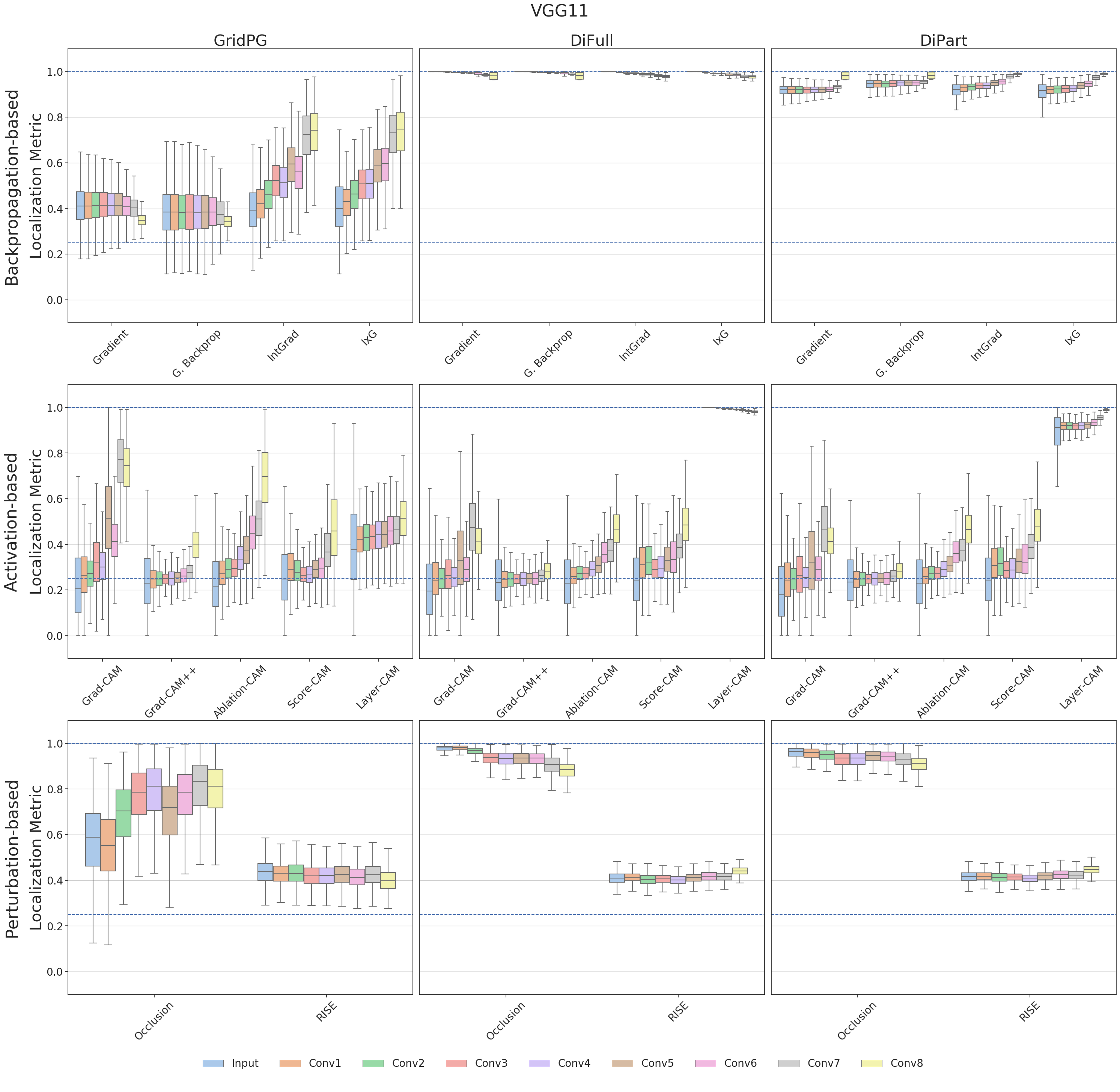}
    \caption{\textbf{Quantitative Results for VGG11 across all convolutional layers}. We evaluate the localization scores each attribution method at the input and each convolutional layer of VGG11, on each of \gridpg, \difull, and \dipart. \emph{Top:} Backpropagation-based methods. \emph{Middle:} Activation-based methods. \emph{Bottom:} Perturbation-based methods. The two horizontal dotted lines mark localization scores of $1.0$ and $0.25$, which correspond to perfect and random localization, respectively. We find that the trends in performance corroborate with those seen across the selected input, middle, and final layers in our experiments.}
    \label{fig:alllayers:vgg11}
\end{figure}

\begin{figure}[H]
    \centering
    \includegraphics[width=\linewidth]{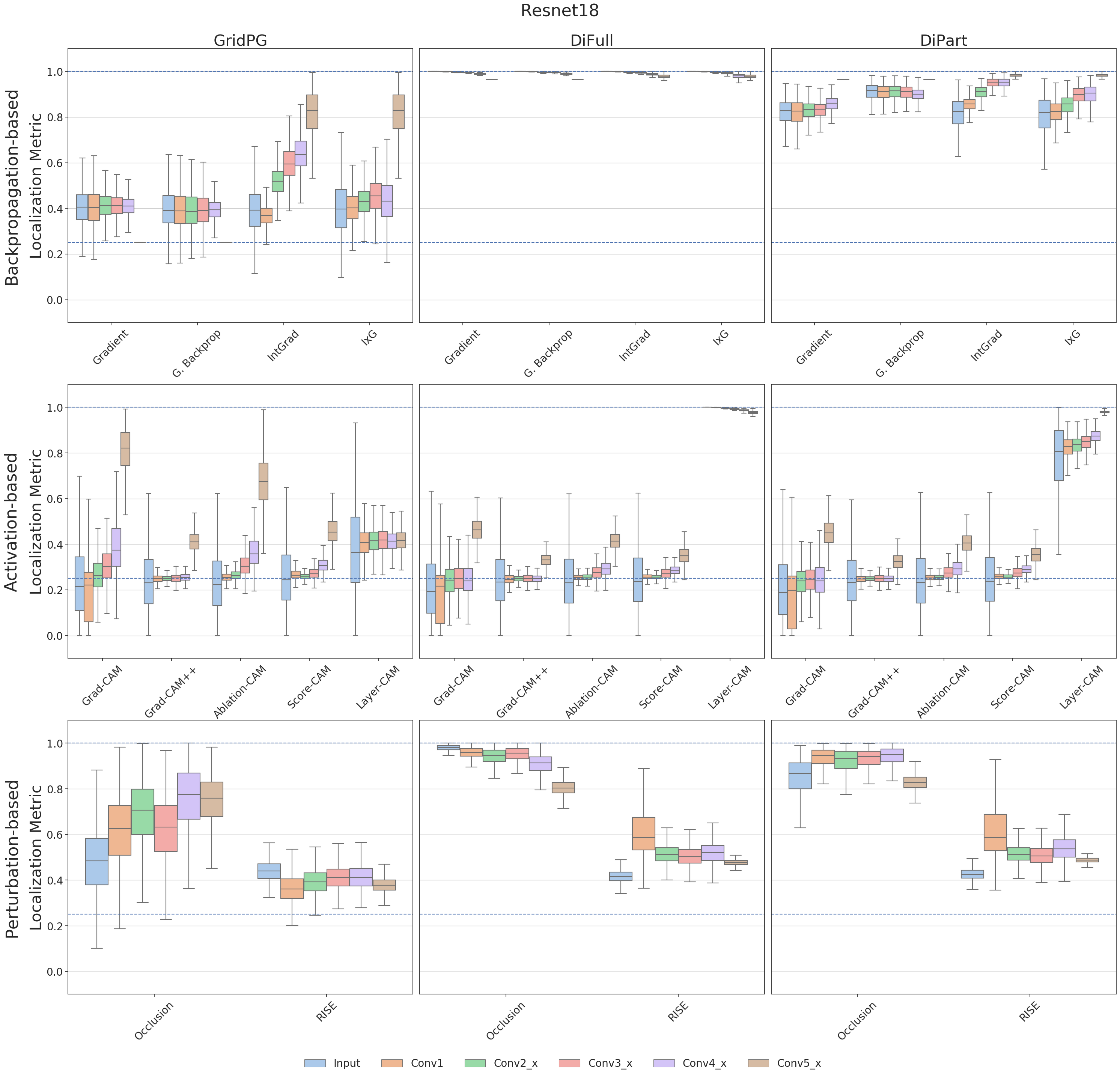}
    \caption{\textbf{Quantitative Results for Resnet18 across all convolutional layer blocks}. We evaluate the localization scores each attribution method at the input and each convolutional layer of Resnet18, on each of \gridpg, \difull, and \dipart. \emph{Top:} Backpropagation-based methods. \emph{Middle:} Activation-based methods. \emph{Bottom:} Perturbation-based methods. The two horizontal dotted lines mark localization scores of $1.0$ and $0.25$, which correspond to perfect and random localization, respectively. We find that the trends in performance corroborate with those seen across the selected input, middle, and final layers in our experiments.}
    \label{fig:alllayers:resnet18}
\end{figure}

For a fair comparison, we evaluated each method at the input, a middle layer, and the final layer of the network. The middle layer was chosen as a representative to visualize the trends in the localization performance across the network. \cref{fig:alllayers:vgg11,fig:alllayers:resnet18} show the results on evaluating at each convolutional layer of VGG11 \cite{simonyan2014very} and each layer block of Resnet18 \cite{he2016deep}. We find that the performance on the remaining layers is consistent with the trend observed from the three chosen layers in our experiments.

\newpage
\section{Computational Cost}\label{sec:cost}
Unlike \gridpg \cite{boehle2021convolutional}, the \difull setting involves passing each grid cell separately through the network. In this section, we compare the computational costs of \gridpg, \difull, and \dipart, and show that it is similar across the three settings. Let the input be in the form of a $n\times n$ grid. Each setting consists of a CNN module, which obtains features from the input, and a classifier module, which provides logits for each cell in the grid using the obtained features. We analyze each of these modules one by one. 

\paragraph{CNN Module:} In \gridpg and \dipart, the entire grid is passed through the CNN module as a single input. On the other hand, in \difull, each grid cell is passed separately. This can be alternatively viewed as stacking each of the $n^2$ grid cells along the batch dimension before passing them through the network. Consequently, the inputs in the \difull setting have their widths and heights scaled by a factor of $\frac{1}{n}$, and the batch size scaled by a factor of $n^2$. Since the operations within the CNN module scale linearly with input size, the computational cost for each grid cell in \difull is $\frac{1}{n^2}$ times the cost for the full grid in \gridpg and \dipart. Since there are $n^2$ such grid cells, the total computational cost for the CNN module of \difull equals that of \gridpg and \dipart.

\paragraph{Classifier Module:} The classifier module in the \difull and \dipart settings consists of $n^2$ classification heads, each of which receives features corresponding to a single grid cell. On the other hand, the \gridpg setting uses a classifier kernel over the composite feature map for the full grid. Let the dimensions of the feature map for a single grid cell be $d\times d$. This implies that in \gridpg, using a stride of 1, the classification kernel slides over $((n-1)d+1)^2$ windows of the input, each of which results in a call to the classifier module. In contrast, in \difull and \dipart, the classifier module is called only $n^2$ times, one for each head. This shows that the computational cost of \difull and \dipart for the classifier module and the pipeline as a whole is at most as much as of \gridpg.

\newpage
\section{Comparison with \smoothgrad}\label{sec:smoothgrad}
\begin{figure}[H]
    \centering
    \includegraphics[width=0.5\linewidth]{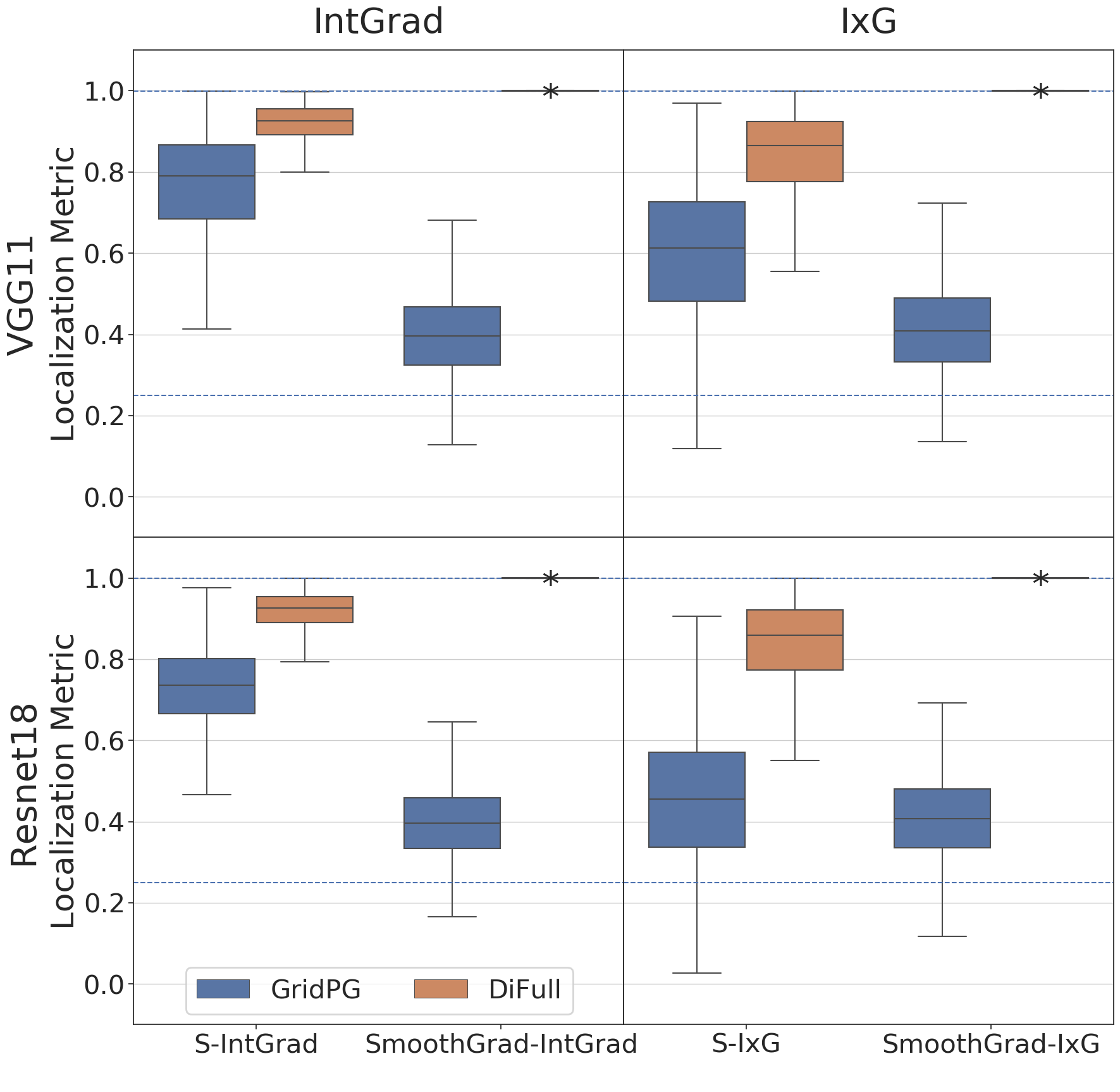}
    \caption{\textbf{Quantitative Results comparing our smoothing with SmoothGrad using VGG11}. We evaluate each attribution method at the input layer. For \sintgrad and \sixg, we use $K=129$. For SmoothGrad, we use the best performing configuration after varying the noise percentage from 1\% to 30\%, and use 15 samples per input. \emph{Top:} Results on VGG11. \emph{Bottom:} Results on Resnet18. We use the ``*'' symbol to show boxes that collapse to a single point, for better readability.}
    \label{fig:smoothgrad}
\end{figure}

\begin{figure}[H]
	\centering
	\includegraphics[width=0.5\linewidth]{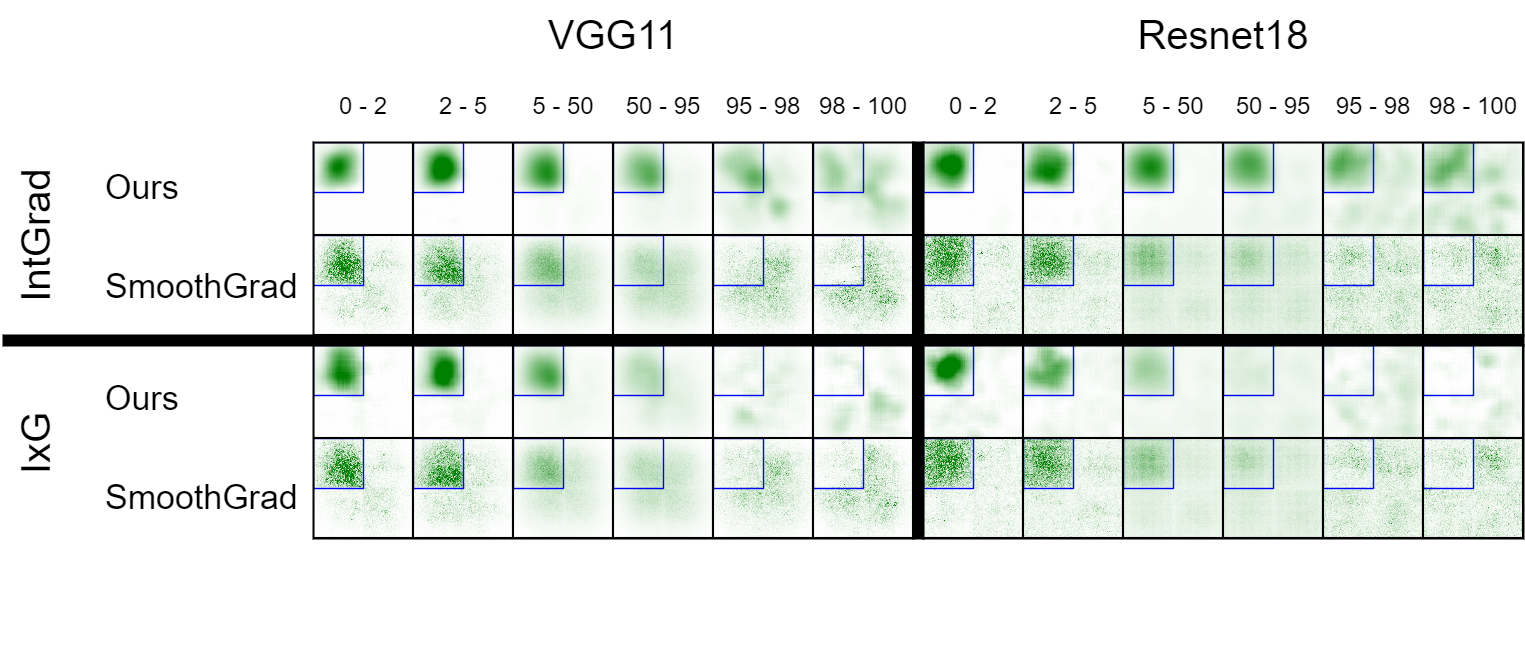}
	\caption{\textbf{\aggatt visualizations of our smoothing (\sintgrad,\sixg\intgrad) compared with SmoothGrad applied on \intgrad and \ixg using VGG11 and Resnet18 on \gridpg.}}
	\label{fig:vis:bins:smoothgrad:gridpg}
\end{figure}

In our work, we find that smoothing \intgrad \cite{sundararajan2017axiomatic} and \ixg \cite{shrikumar2017learning} attributions with a Gaussian kernel can lead to significantly improved localization, particularly for networks without batch normalization layers \cite{ioffe2015batch}. As discussed in \cref{sec:smooth:explain}, we believe this to be because smoothing summarizes the effect of inputs in a local window around each pixel to the output logit, and reduces noisiness of attributions. Prior approaches to address noise in attributions include SmoothGrad \cite{smilkov2017smoothgrad}, which involves adding Gaussian noise to an input and averaging over attributions from several samples. Here, we compare our smoothing with that of SmoothGrad. \cref{fig:smoothgrad} shows that our methods (\sintgrad, \sixg) show significantly better \gridpg \cite{boehle2021convolutional} localization than SmoothGrad on \intgrad and \ixg, except in the case of \ixg with Resnet18 \cite{he2016deep}, where our smoothing does not improve localization likely due to the presence of batch normalization layers. The scores on \difull decrease to an extent since our Gaussian smoothing allows attributions to ``leak'' to neighbouring grid cells. These results are corroborated by \aggatt visualizations in \cref{fig:vis:bins:smoothgrad:gridpg}. We also note that \smoothgrad requires significantly higher computational cost than our approach, as attributions need to be generated for several noisy samples of each input, and is also sensitive to the choice of hyperparameters such as the noise percentage and the number of samples.


\newpage
\section{Implementation Details}\label{sec:impldetails}
\subsection{Dataset}\label{sec:impldetails:dataset}
As described in the paper (Sec. 4), we obtain 2,000 attributions for each attribution method on each of \gridpg \cite{boehle2021convolutional}, \difull, and \dipart, using inputs consisting of four subimages arranged in $2\times2$ grids. For \gridpg, since we evaluate on all four subimages, we do this by constructing 500 grid images after randomly sampling 2,000 images from the validation set. Each grid image contains subimages from four distinct classes. On the other hand, for \difull and \dipart, we place images of the same class at the top-left and bottom-right corners to test whether an attribution  method  simply  highlights  class-related  features, irrespective of them being used by the model. Therefore, we evaluate only on these two grid locations. In order to obtain 2,000 attributions as with \gridpg, for these two settings, we construct 1,000 grid images by randomly sampling 4,000 images from the validation set.

\subsection{Models and Attribution Methods}
We implement our settings using PyTorch \cite{pytorch}, and use pretrained VGG11 \cite{simonyan2014very} and Resnet18 \cite{he2016deep} models from Torchvision \cite{pytorch}. We use implementations from the Captum library \cite{kokhlikyan2020captum} for \grad \cite{simonyan2013deep}, \gb \cite{springenberg2014striving}, \intgrad \cite{sundararajan2017axiomatic}, and \ixg \cite{shrikumar2017learning}, and from \cite{boehle2021convolutional} for \occ \cite{zeiler2013visualizing} and \rise \cite{petsiuk2018rise}. For \grad and \gb, the absolute value of the attributions are used. All attributions across methods are summed along the channel dimensions before evaluation.

\occ involves sliding an occlusion kernel of size $K$ with stride $s$ over the image. As the spatial dimensions of the feature maps decreases from the input to the final layer, we select different values of $K$ and $s$ for each layer. In our experiments, we use $K=16,s=8$ for the input, and $K=5,s=2$ for the middle and final layers.

\rise generates attributions by occluding the image using several randomly generated masks and weighing them based on the change in the output class confidence. In our experiments, we use $M=1000$ masks. We use fewer masks than \cite{petsiuk2018rise} to offset the increased computational cost from using $448\times448$ images, but found similar results from a subset of experiments with $M=6000$.

\subsection{Localization Metric}
In our quantitative evaluation, use the same formulation for the localization score as proposed in \gridpg (Sec 3.1.1 in the paper).
Let $A^+(p)$ refer to the positive attribution given to the $p^{th}$ pixel. The localization score for the subimage $x_i$ is given by:
\begin{equation}\label{eq:localization}
    L_i = \frac{\sum_{p\in x_i} A^+(p)}{\sum_{j=1}^{n^2}\sum_{p\in x_j} A^+(p)} 
\end{equation}
However, $L_i$ is undefined when the denominator in \cref{eq:localization} is zero, \ie, $\sum_{j=1}^{n^2}\sum_{p\in x_j} A^+(p)=0$. This can happen, for instance, when all attributions for an input are negative. To handle such cases, we set $L_i=0$ in our evaluation whenever the denominator is zero.

\subsection{\aggatt Visualizations}
To generate our \aggatt visualizations, we sort attribution maps in the descending order of the localization score and bin them into percentile ranges to obtain aggregate attribution maps (Sec 3.2 in the paper). However, we observe that when evaluating on \difull, the backpropagation-based attribution methods show perfect localization (Sec 5.1 in the paper), and all attributions share the same localization score. In this scenario, and in all other instances when two attributions have the same localization score, we break the tie by favouring maps that have stronger attributions in the target grid cell. We do this by ordering attributions with the same localization score in the descending order of the sum of attributions within the target grid cell, \ie, the numerator in \cref{eq:localization}.

Further, when producing the aggregate maps, we normalize the aggregate attributions using a common normalizing factor for each method. This is done to accurately reflect the strength of the average attributions across bins for a particular method.


\newpage
\section{Evaluation on CIFAR10}\label{sec:cifar}
In addition to ImageNet \cite{imagenet}, we also evaluate using our settings on CIFAR10 \cite{krizhevsky2009learning}. In this section, we present these results, and find similar trends in performance as on ImageNet. We first describe the experimental setup (\cref{sec:cifar:setup}) used, and then show the quantitative results on \gridpg \cite{boehle2021convolutional}, \difull, and \dipart (\cref{sec:cifar:quantitative}) and some qualitative results using \aggatt (\cref{sec:cifar:qualitative}).

\subsection{Experimental Setup}\label{sec:cifar:setup}
\myparagraph{Network Architecture:}
We use a modified version of the VGG11 \cite{simonyan2014very} architecture, with the last two convolutional layers removed. Since the CIFAR10 inputs have smaller dimensions ($32\times32$) than Imagenet ($224\times224$), using all the convolutional layers results in activations with very small spatial dimensions, which makes it difficult to apply attribution methods at the final layer. After removing the last two convolutional layers, we obtain activations at the new final layer with dimensions $4\times4$ before pooling. We then perform our evaluation at the input (Inp), middle layer (Conv3) and the final layer (Conv6).

\myparagraph{Data:}
We construct grid datasets consisting of $2\times2$ and $3\times3$ grids using images from the validation set classified correctly by the network with a confidence of at least $0.99$. We obtain 4,000 (resp. 4,500) attributions for each method from the $2\times2$ (resp. $3\times3$) grid datasets respectively. As with ImageNet (\cref{sec:impldetails:dataset}), we evaluate on all grid cells for \gridpg and only at the top-left and bottom-right corners on \difull and \dipart. To obtain an equivalent 4,000 (resp. 4,500) attributions from using just the corners on \difull and \dipart, we randomly sample 8,000 (resp. 20,250) images for the $2\times2$ ($3\times3$) grid datasets, and construct 2,000 (2,250) composite images. Note that the CIFAR10 validation set only has a total of 10,000 images. Since we only evaluate at the two corners, we allow subimages at other grid cells to repeat across multiple composite images. However, no two subimages are identical within the same composite image.

\subsection{Quantitative Evaluation on \gridpg, \difull, and \dipart}\label{sec:cifar:quantitative}
\begin{figure}[H]
    \centering
    \includegraphics[width=0.9\linewidth]{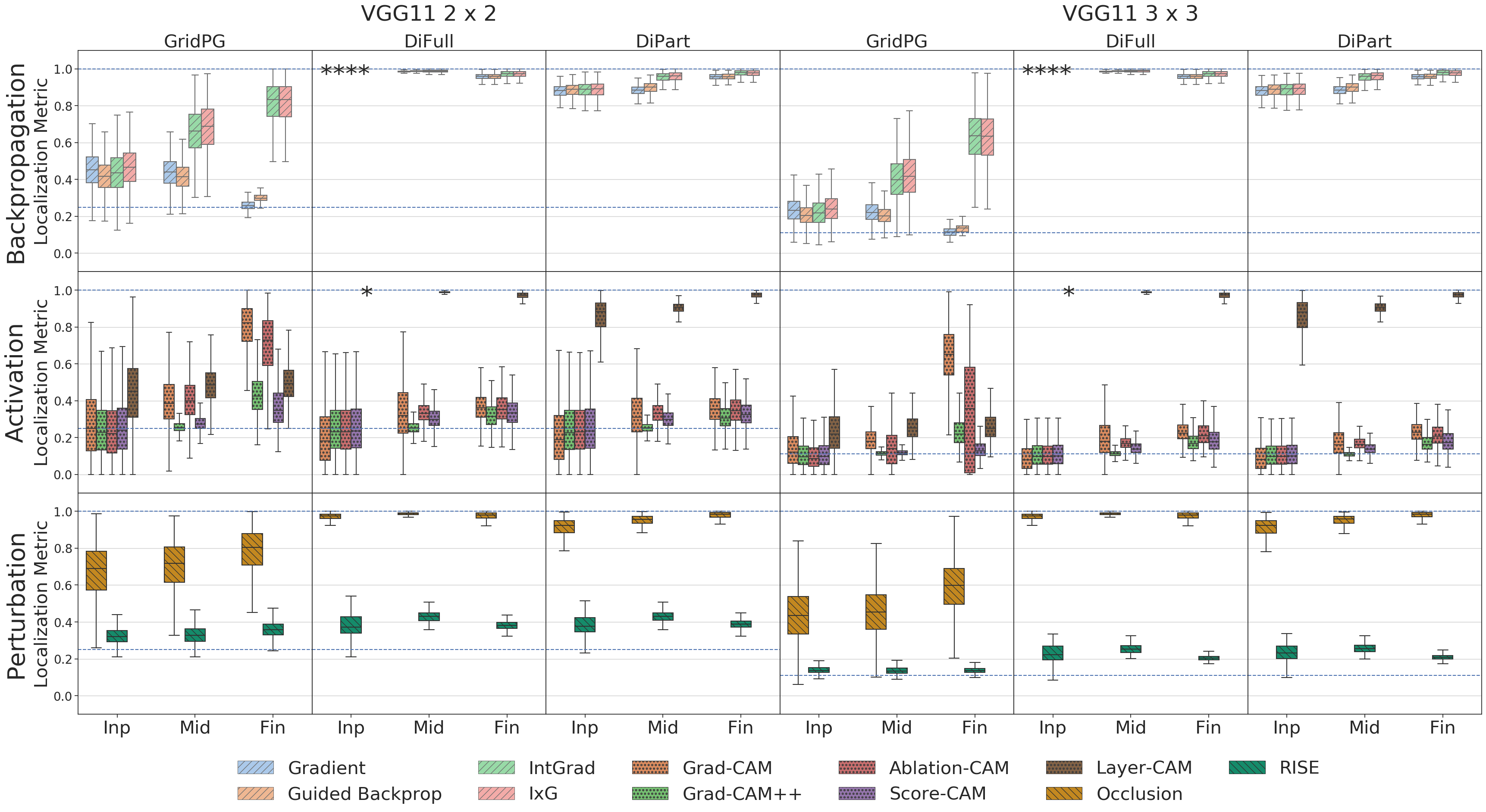}
    \caption{\textbf{Quantitative Results on CIFAR10 using VGG11}. We evaluate each attribution method at the input (Inp), middle (Mid), and final (Fin) convolutional layers, on each of \gridpg, \difull, and \dipart using $2\times2$ (left) and $3\times3$ (right) grids. \emph{Top:} Results on backpropagation-based methods. \emph{Middle:} Results on activation-based methods. \emph{Bottom:} Results on perturbation-based methods. The two horizontal dotted lines mark localization scores that correspond to perfect and random localization, respectively, which equal scores of $0.25$ and $0.11$ respectively for $2\times2$ and $3\times3$ grids. We use the ``*'' symbol to show boxes that collapse to a single point, for better readability.}
    \label{fig:cifar:quantitative}
\end{figure}

The results of the quantitative evaluation can be found in \cref{fig:cifar:quantitative} for both $2\times2$ grids (left) and $3\times3$ grids (right). We observe that all methods perform similarly as on ImageNet (\cref{fig:imagenet:quantitative}). Since localizing on $3\times3$ grids poses a more challenging task, we observe generally poorer performance across all methods on that setting.

\subsection{Qualitative Results using \aggatt}\label{sec:cifar:qualitative}
\begin{figure}[H]
    \centering
    \includegraphics[width=0.5\linewidth]{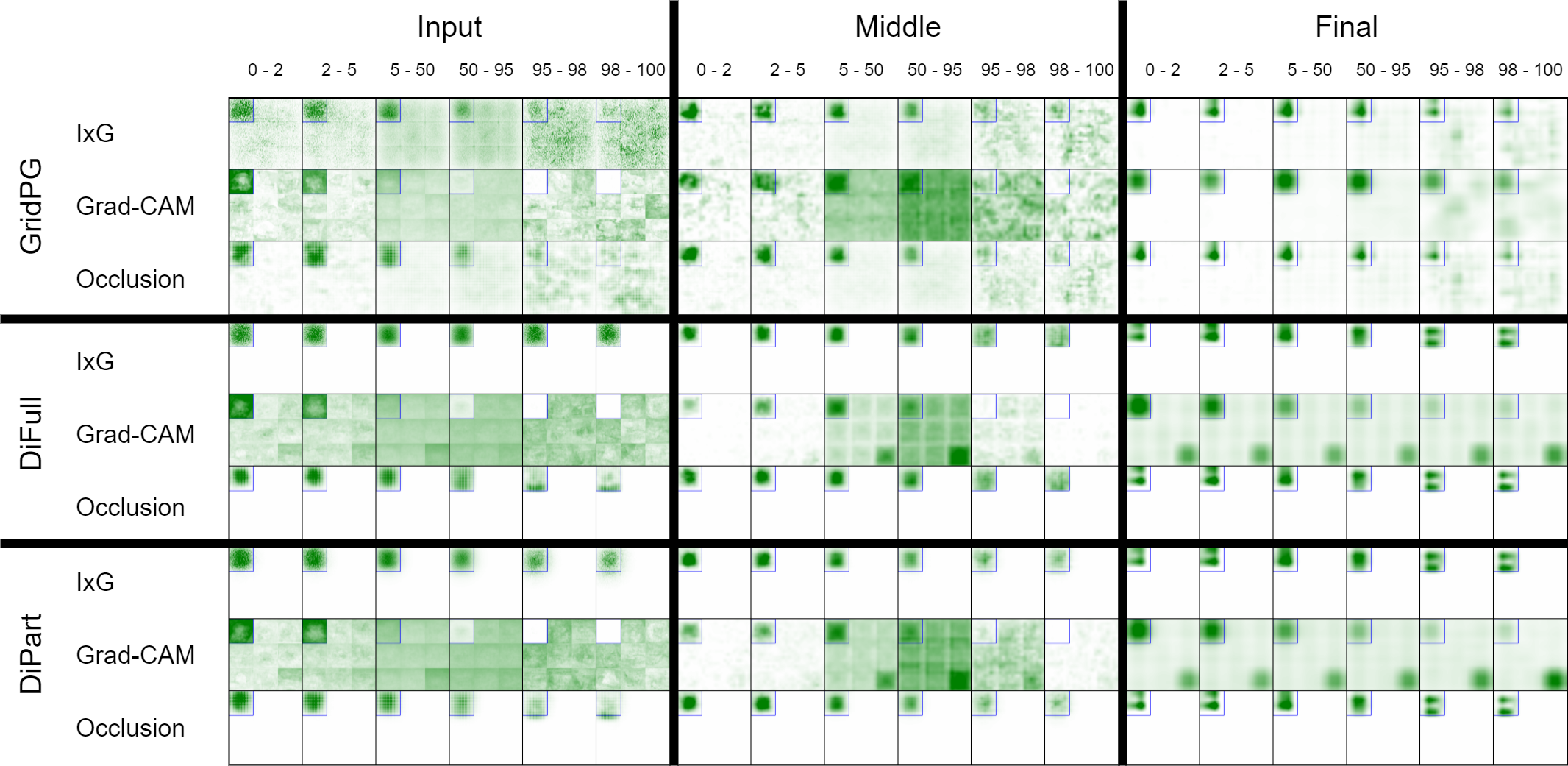}
    \caption{\textbf{\aggatt Evaluation on \gridpg for all methods at the input, middle, and final layers using VGG11 with the $3\times3$ grid dataset.}}
    \label{fig:vis:cifar:bins}
\end{figure}

\begin{figure}[H]
    \centering
    \includegraphics[width=0.6\linewidth]{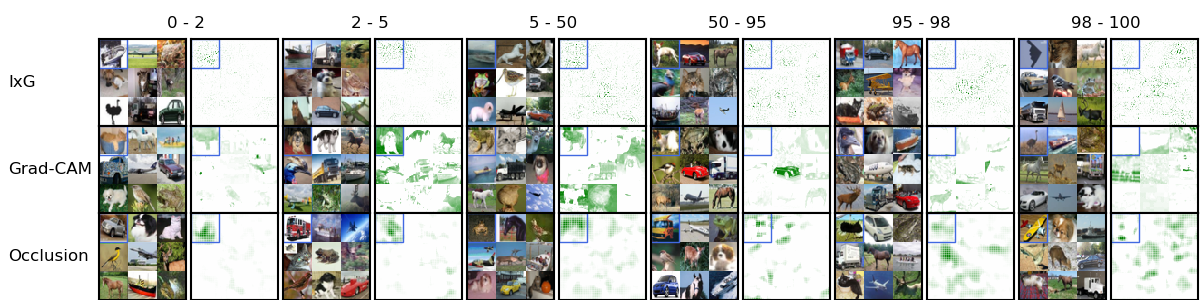}
    \caption{\textbf{Examples from each \aggatt bin for each method at the input layer on \gridpg using VGG11 on $3\times3$ grids.} From each bin, the image and its attribution at the median position are shown.}
    \label{fig:vis:cifar:examples:gridpg:vgg11:input}
\end{figure}

\begin{figure}[H]
    \centering
    \includegraphics[width=0.6\linewidth]{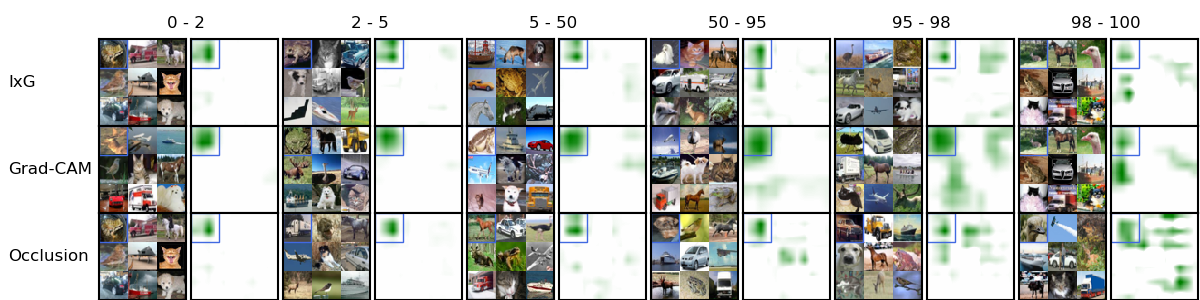}
    \caption{\textbf{Examples from each \aggatt bin for each method at the final layer on \gridpg using VGG11 on $3\times3$ grids.} From each bin, the image and its attribution at the median position are shown.}
    \label{fig:vis:cifar:examples:gridpg:vgg11:final}
\end{figure}

In \cref{fig:vis:cifar:bins}, we show \aggatt evaluations on $3\times3$ grids for a method each from the set of backpropagation-based (\ixg \cite{shrikumar2017learning}), activation-based (\gradcam \cite{selvaraju2017grad}), and perturbation-based (\occ \cite{zeiler2013visualizing}) methods. Further, we show examples of attributions at the input and final layer on \gridpg for these methods (\cref{fig:vis:cifar:examples:gridpg:vgg11:input,fig:vis:cifar:examples:gridpg:vgg11:final}). We see that these show similar trends in their performance as on ImageNet (\cref{sec:additional:aggatt}). 
\newpage

{\small
\bibliographystyle{ieee_fullname}
\bibliography{references}
}